\documentclass{article}

\PassOptionsToPackage{numbers, compress}{natbib}
\usepackage[eandd, final]{neurips_2026}

\usepackage{graphicx}
\usepackage{booktabs}
\usepackage{tcolorbox}
\usepackage{tabularx} 
\usepackage{listings}

\usepackage{multirow}
\usepackage{tcolorbox}
\newcommand{\titletext}{Toward Interactive Understanding of Code APIs}
\usepackage{amsmath,amsfonts,bm}

\newcommand{\benchmarkshort}{{\bf PAU}}
\newcommand{\benchmark}{{\textbf{P}ython \textbf{A}PI \textbf{U}nderstanding}}

\def\eqref#1{equation~\ref{#1}}

\def\1{\bm{1}}

\DeclareMathAlphabet{\mathsfit}{\encodingdefault}{\sfdefault}{m}{sl}
\SetMathAlphabet{\mathsfit}{bold}{\encodingdefault}{\sfdefault}{bx}{n}

\usepackage[utf8]{inputenc} 
\usepackage[T1]{fontenc}    
\usepackage[colorlinks=true,allcolors=black]{hyperref}\usepackage{url}            
\usepackage{booktabs}       
\usepackage{amsfonts}       
\usepackage{nicefrac}       
\usepackage{microtype}      
\usepackage{xcolor}         
\usepackage{color-edits}
\addauthor[Jon]{jon}{blue}
\addauthor[DJ]{dj}{purple}
\title{\titletext}

\author{%
  Dhananjay Ashok \\  
Information Sciences Institute \\
University of Southern California\\
  \texttt{ashokd@usc.edu} \\
  \And
  Jesse Thomason \\
  University of Southern California \\
  \texttt{jessetho@usc.edu} \\
  \And
  Jonathan May \\
  Information Sciences Institute \\
  University of Southern California\\
  \texttt{jonmay@isi.edu} \\
}

\begin{document}

\maketitle

\begin{abstract}
Empowered by advances in Language Model agents, systems have made substantial strides in code generation and understanding. However, these approaches often rely on read access to the relevant code, an assumption which does not hold when dealing with external APIs. In this work, we introduce the \benchmarkshort{} (\benchmark{}) benchmark, where we provide models with black-box, API-level access to code snippets. Models must query the API with exploratory inputs and draw insights from the resulting outputs, with the goal of describing the snippet's true functionality. By treating the code snippets as external tools that must be understood via interaction alone, \benchmarkshort{} studies the more general problem of unsupervised tool understanding, specifically for tools implemented as Python methods. Despite recent progress in coding agents, even frontier models struggle to achieve high performance on \benchmarkshort{}, with the best model (Claude-4-Opus) failing to understand over 45\% of the \benchmarkshort{} test set. An investigation into the common error modes reveals that models are overconfident; they often overrate the quality of their current hypothesis, leading to insufficient exploration and premature termination. Finally, we take inspiration from the Asymmetric Actor Critic (AAC) paradigm, frequently used in robot learning, to post-train models for interactive code understanding. Models trained with AAC conduct more active exploration of the APIs, with an AAC-tuned Qwen3-8B model matching the performance of GPT-5-mini.\footnote{Code and data available at: \url{https://github.com/DhananjayAshok/PAU-Benchmark}}

\end{abstract}

\section{Introduction}
\label{sec:introduction}
With powerful Language Models (LMs) at their core~\citep{brown2020language}, the latest generation of coding-agents~\citep{anthropic2025claudecode} set new standards in automated software development and code understanding~\citep{jain2025livecodebench}. Existing approaches to evaluation often provide the model with complete, read access to the code under investigation~\citep{wang-etal-2025-coderag, yang2023intercode}. For instance, ~\citet{jimenez2024swebench} provide models with read access over GitHub repositories, and task them with resolving an issue. However, programmers often find themselves conducting \textit{exploratory programming}~\citep{kery2017exploring} where they must rely on external or proprietary Application Programming Interfaces (APIs), which obscure their internal code~\citep{jacobson2012apis}. The critical role of such APIs in software workflows necessitates the ability to understand code artefacts by studying their behaviour, as opposed to inspecting their source code~\citep{mammadov2024learning}. For instance, while LM agents can identify zero-day vulnerabilities in open-sourced kernels by inspecting their code~\citep{Brumley2026, MicrosoftCopyFail2026}, detecting vulnerabilities in closed kernels would require a more interactive approach. Vital tasks in this setting, such as black-box testing of APIs~\citep{alonso2023agora} and reverse engineering~\citep {yang2026apisensor}, remain challenging for more traditional, formal approaches~\citep{woodcock2009formal}, which rely on descriptive specifications for the API under exploration~\citep{neider2024formal}. 

\begin{figure}
    \centering
    \includegraphics[width=\linewidth]{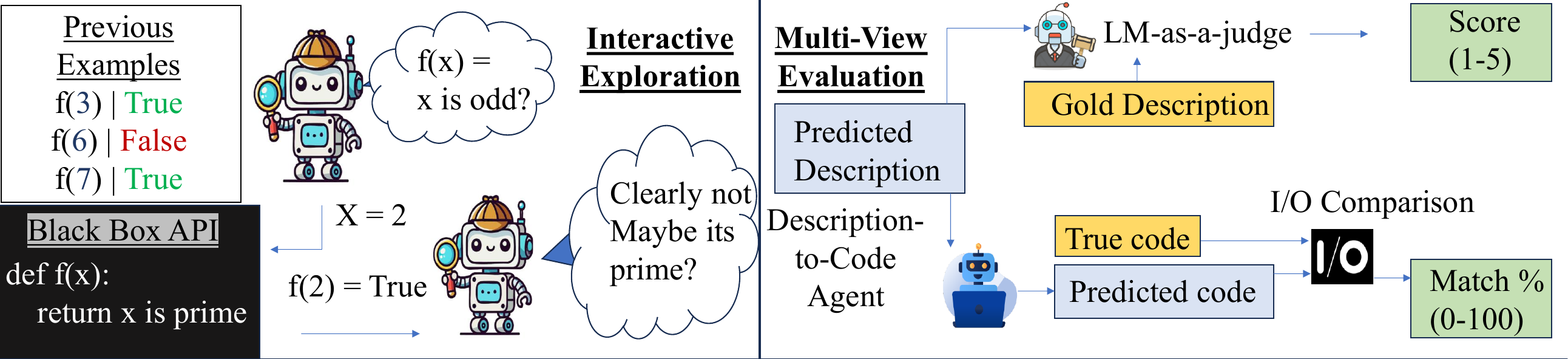}
    \caption{The \benchmarkshort{} benchmark for interactive understanding of code APIs. Models are given (left) black box access to a Python method and must explore it enough to produce a description of its functionality. Given a predicted description, we evaluate it from two different perspectives. Using the gold description of the method's functionality, we score the model's prediction from 1-5 (inclusive) using an LM-as-a-judge (top-right). Additionally, we use a fixed code-writing model to convert the predicted description to a code snippet. Using a test suite of inputs, we assess the percentage of inputs for which the output of the snippet matches that of the true code (bottom-right).}
    \label{fig:pau}
\end{figure}

LMs, with their capacity to reason over their input contexts and interact with APIs, hold promise in such under-specified domains. Despite this potential, the ability of models to effectively explore and understand code through interaction alone is not well understood~\citep{wang-etal-2025-exploracoder, yin2026investigating}. To bridge this gap, we construct \benchmarkshort{} (Python API Understanding) (Figure~\ref{fig:pau}), a benchmark for interactive understanding of code APIs. Specifically, models are provided with black-box, API access to a Python method (i.e. can query with an input and observe an output, but cannot access the internal function body). The models are tasked with understanding the API enough to accurately describe the method's true functionality in words. Strong performance requires a model to effectively explore the API with meaningfully different inputs and draw insights from the resulting outputs.

Despite impressive performance on standard code-understanding benchmarks~\citep{jimenez2024swebench} even frontier models struggle on \benchmarkshort{}, with every model (including Gemini-3-Flash~\citep{team2024gemini}, GPT-4o~\citep{openai2024gpt4o}, Claude-4-Opus~\citep{anthropic2026opus47} and DeepSeek-V3~\citep{DeepSeekAI2025DeepSeekR1IR}) failing to understand over 45\% of the available methods. Worryingly, providing two static input-output examples instead of interactive access to the API often leads to similar performance. This lack of improvement when models are allowed API access suggests that they fail to progressively improve their understanding over multiple rounds of interaction. 

An inspection of the errors reveals a consistent failure mode related to the ability of agents to evaluate their own progress. Models are often overconfident and conduct shallow exploration. When evaluating the quality of their hypotheses, they often prematurely decide that they have fully understood the API, opting to terminate without a thorough exploration of the API's functionality. We hypothesise this problem is related to the deeper phenomena of sycophancy~\citep{fanous2025syceval} and inflated self-judgement~\citep{lu-etal-2025-llm}. 

We establish (Section~\ref{sec:solution}) that due to a lack of annotated exploration data and systematic unit tests to provide verifiable reward signals, the traditional approaches of Instruction-tuning~\citep{raffel2020exploring} and Reinforcement Learning from Verifiable Rewards (RLVR)~\citep{lambert2025tulu} are inapplicable to our setting. For a solution, we take inspiration from robot learning~\citep{siciliano2008springer} and adapt the asymmetric actor-critic (AAC) paradigm~\citep{pinto2017asymmetric} to tune Language Models~\citep{zheng2023secrets} for interactive code understanding. This paradigm is a reinforcement learning (RL)~\citep{kaelbling1996reinforcement} framework, where the \textbf{critic} model generates high-quality guidance using privileged information that would not be accessible during agent deployment. 

Specifically, we use RL-algorithms (e.g. PPO~\citep{schulman2017proximal}) to fine-tune a model for interactive code understanding. At each step of RL-training, the agent attempts to interact with and describe its understanding of the API. Its explorations and working hypotheses are evaluated for rewards by an LM-as-a-critic. Different from distillation-based paradigms, which rely on a more powerful judge model to generate useful rewards ~\citep{huang2022knowledge, zheng2023judging}, our LM-as-a-critic uses the \textbf{same} backbone as the agent being trained and instead relies on privileged information~\citep{cai2024provable} to produce valid step-level rewards. Specifically, the critic model is provided with the true description of the method that the actor is attempting to understand, allowing it to generate more precise rewards for exploration trajectories. Across multiple base models with varying scales, we show that while supervised fine-tuning has no impact on model performance, the AAC paradigm consistently improves their ability to explore an API and understand its functionality. An AAC-tuned Qwen3-8B model~\citep{yang2025qwen3} matches the performance of the closed-source GPT-5-mini model, demonstrating the power of the approach.

We summarize our contributions as follows:
\begin{enumerate}
    \item The introduction of the \benchmarkshort{} benchmark, which evaluates an agent's ability to understand code through interaction with code APIs as opposed to evaluating their code-reading abilities. We are the first to demonstrate that while frontier models excel at understanding written code, they fail at understanding code artefacts through interaction alone. 
    \item An analysis of the failure modes of LM-based, interactive agents, showcasing consistent overconfidence and premature termination. These failure patterns persist across LMs of various scales and training paradigms, suggesting a limitation of current LM-based agents. 
    \item The translation of the Asymmetric Actor Critic paradigm from robot learning settings to Language Model-based RL. We are the first to use AAC to RL-tune LMs, and the approach's success opens the way for its application in other domains.
\end{enumerate}

\section{Python API Understanding Benchmark}
\label{sec:benchmark}

\benchmarkshort{} is a collection of 3.9K train and 741 test instances of the form: $<f, d, E_{\text{context}}, E_{\text{test}}>$.

\begin{itemize}
    \item $f$: Python code implementing a method that takes input arguments and returns an output value. The declaration of every method follows the form: `def test\_func(arg0, ...)' to prevent the header from revealing information regarding its functionality. All methods are stateless, i.e. each call to $f$ is processed independently, with no dependence on previous calls.
    \item $d$: text of a gold description that explains the functionality of $f$. 
    \item $E_{\text{context}} = \{(i_1, o_1), (i_2, o_2)\}$: exactly two valid input-output pairs for contextualization. These examples reveal some functionality of $f$, but do not showcase its full workings. 
    \item $E_\text{test}=\{i_3, \ldots i_n\}$: a suite of test inputs which characterizes the functionality of $f$. Given a mapping $\hat{f}$, if $\forall i\in E_\text{test}, \hat{f}(i)=f(i)$ then we consider $\hat{f}$ to be functionally identical to $f$. 
\end{itemize}

 During evaluation, the model starts with the header of $f$ and $E_{\text{context}}$ (to help contextualize the possible space of input arguments). It is also given black-box, API access to $f$, accepting arbitrary Python tuples as input arguments and providing the returned output in response. The goal of the model is to write a description $\hat{d}$ of the functionality of $f$. 

 \noindent\textbf{Evaluation:} We adopt a multi-view perspective using reproducible open-weight models~\citep{atil-etal-2025-non} to robustly evaluate the extent to which a predicted description captures the true functionality of $f$. 

\underline{LM-as-a-judge}: Classic approaches to comparing strings (e.g. BLEU, etc~\citep{schmidtova-etal-2024-automatic-metrics}) are problematic, as descriptions of methods with opposite functionalities may have high lexical overlap (e.g. `True if arg0 is even' v.s. `True if arg0 is odd'). We use an LM-as-a-judge~\citep{gu2024survey}(Llama3-70B~\citep{grattafiori2024llama}) to score the extent to which the predicted description reflects the true functionality on a Likert scale of 1-5~\citep{likert1932technique}. We conduct validation tests, confirming that the rankings between models are unaffected by the specific choice of prompt used, and the scores of our judge model show high agreement with both human judgements and labels from more powerful models (see Appendix~\ref{sec:evalrobust} for details). Additionally, when we re-score the top four models (GPT-5-mini, Gemini-3-Flash, Claude-4-Sonnet, Claude-4-Opus) using frontier judges from two different providers (Gemini-3-Flash and Claude-4-Sonnet), the model ranking is unchanged, and scores drop only marginally (by 0.06 on average). Since description scores are an aggregation of LM-as-a-judge ratings, we caution against interpreting their absolute magnitude outside of the extremes, and view them primarily as a tool for ordinal comparison between models.

\underline{Description-to-Code}: An accurate and complete description can be used to reverse engineer the code of the method~\citep{chen2021evaluating}. We use a dedicated coding model (Qwen3-Coder-30B-A3B~\citep{yang2025qwen3}) to translate $\hat{d}$ into a Python method $\hat{f}$ and measure the `code match' score, i.e., the extent to which it accurately reproduces the behavior of $f$ on the test suite: $\frac{1}{|E_\text{test}|}\sum_{i\sim E_\text{test}}\mathbb{I}[f(i)=\hat{f}(i)]$. To verify the strength of our coding model, we compare its performance to GPT-4o when writing code for the \textbf{gold descriptions}. Qwen3-Coder-30B-A3B marginally outperforms GPT-4o(average code match 72.96\% vs 72.06\%), confirming its suitability as a code-writing model. The imperfect performance of the code-writing model shows that the description-to-code task is still unsolved for the methods in \benchmarkshort{}, and that 100\% code match is not yet attainable. 

Given the underspecified nature of assessing the semantic equivalence of code methods or their descriptions~\citep{brooks1987no, zeng2025veriequivbench}, neither approach guarantees a perfect metric. However, their combination allows us to isolate strong and clear signals of differences between models and methods.

\noindent\textbf{Benchmark Construction:} We source the methods and gold descriptions from existing description-to-code benchmarks. For the training set, we source from Code-Alpaca~\citep{codealpaca}, and for the test set, we use HumanEval~\citep{chen2021evaluating}, MBPP~\citep{austin2021program} and Crux-Eval~\citep{gu2024cruxeval}. We use string parsing and manipulation to rename the function declaration and input arguments. Finally, we prompt an LM (GPT-4o) to provide an exhaustive set of inputs that would characterize the performance of the function. We then perform multiple rounds of filtering: we remove methods that do not have at least five different test inputs and remove methods where all test inputs produce the same output. We adopt the industry gold standard for verifying that our test suite fully characterizes the functionality of each method by filtering based on \textbf{statement} and \textbf{branch coverage}~\citep{Goodenough1975TowardAT, zhu1997software}. After filtering, the average statement/branch coverage across the test methods is over 99\%, confirming that the test suites cover the functionality of the method. For exact details (exact prompts, tooling used, testbed statistics, etc.), see Appendix~\ref{sec:appendix_benchmark}. 

\section{Evaluating Models for Interactive Function Discovery }
\label{sec:results}

We evaluate a range of models on \benchmarkshort{}, with different underlying training paradigms:

\noindent\textbf{Instruction-tuned:} Standard LMs that have been trained via pretraining, then instruction-tuning~\citep{brown2020language}, and finally RL-tuning with human feedback~\citep{zheng2023secrets}. This includes the Qwen3 model family (1.7B, 8B) and the Llama3 model family (1B, 8B, 70B). 

\noindent\textbf{Coding Models:} LMs that have been explicitly trained for code-generation and understanding. Specifically, we use Qwen3-Coder-30B-A3B-Instruct and Qwen3-Coder-Next, a pair of strong mixture-of-experts models trained on data that includes instances of API calling. 

\noindent\textbf{Reasoning Models:} LMs tuned with RLVR, often with a focus on mathematical reasoning, code writing and scientific understanding~\citep{DeepSeekAI2025DeepSeekR1IR}. In particular, we benchmark the most capable open-weight models, DeepSeekV3 and Gemma4-31B~\citep{team2024gemma}. 

\noindent\textbf{Frontier Models:} State-of-the-art models from closed source providers. Information on their training regimes and scale is scarce. We use powerful models from OpenAI (GPT-4o-mini, GPT-4o, GPT-5.4-mini), Anthropic (Claude-4-Sonnet, Claude-4-Opus) and Google (Gemini-3-Flash). 

We evaluate each model in two different modes:

\noindent\textbf{In-context:} An \textbf{interaction-free} setting where we do not provide models with API access to $f$. Models must use only the two examples in $E_{\text{context}}$ to predict the functionality of $f$. We use this mode to quantify the added benefit of providing models with interactive API access. 

\noindent\textbf{Interactive:} For the standard evaluation mode in \benchmarkshort{}, we implement a fixed orchestration agentic workflow, following a step-wise ReACT paradigm~\citep{yao2022react}. The model maintains a running hypothesis regarding the method's functionality and records the list of all observed input-output examples. The orchestration constrains the agent to alternate between three distinct phases (Figure~\ref{fig:interactive_figure}):
\begin{figure}
    \centering
    \includegraphics[width=\linewidth]{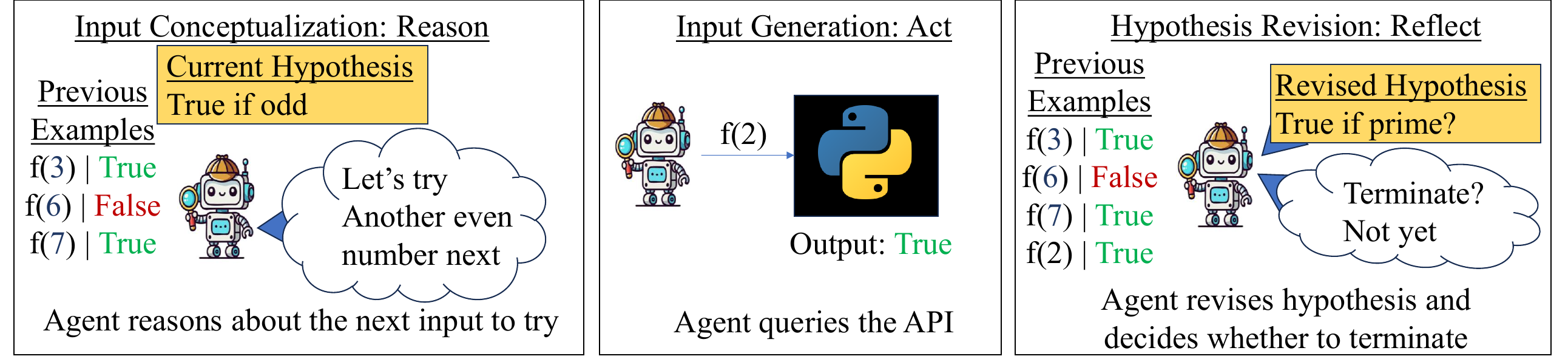}  
    \caption{Orchestration structure for API understanding. Agents engage in a reason-act-reflect loop.}
    \label{fig:interactive_figure}
\end{figure}

\begin{enumerate}
    \item Next input conceptualization: The model is provided with its current hypothesis, the method header, and the observed input-output examples (initially $E_{\text{context}}$). The model reasons as to the properties of the input to try next, with the goal of confirming or rejecting the hypothesis. 
    \item Next input generation: Given all of the input information for the previous step (hypothesis, examples, header) and the output of the previous step, the model is prompted to generate a precise Python tuple that can be supplied as input to $f$. We obtain the return value or error using $f$ and add these to the list of seen input-output examples
    \item Hypothesis revision: Given the information for the previous steps, as well as the output to the previously predicted input to the method, the model is prompted to revise its hypothesis if needed, and if confident in its hypothesis, decide to terminate its exploration procedure. If the model does not decide to terminate (and the maximum number of iterations is not exceeded), then the model returns to step 1 with the revised hypothesis and larger input-output list. 
\end{enumerate}

This mode allows models to act as agents, interacting with the API until they autonomously decide that they are confident in their current answer. Our choice of fixed orchestration is driven by initial results, with models struggling to properly query APIs when using open-ended agentic generation. 

To ensure that the performance of this approach is representative of model capabilities as a whole, we experiment with several alternate baselines in the Appendix. We find that performance is robust to the choice of prompts and does not improve when prompts include software testing-specific hints~\citep{myers2004art} (e.g. advocating for boundary-value analysis, query diversification, or providing counterexample search strategies). We also verify that alternate interaction structures and a RAG-based approach using the \benchmarkshort{} training set achieves similar performance, establishing that the \textbf{interactive} setting is a sound estimate of model ability. For exact prompts and alternate baselines explored, see Appendix~\ref{sec:appendix_interactive}.

\begin{figure}
    \centering
    \includegraphics[width=0.49\linewidth]{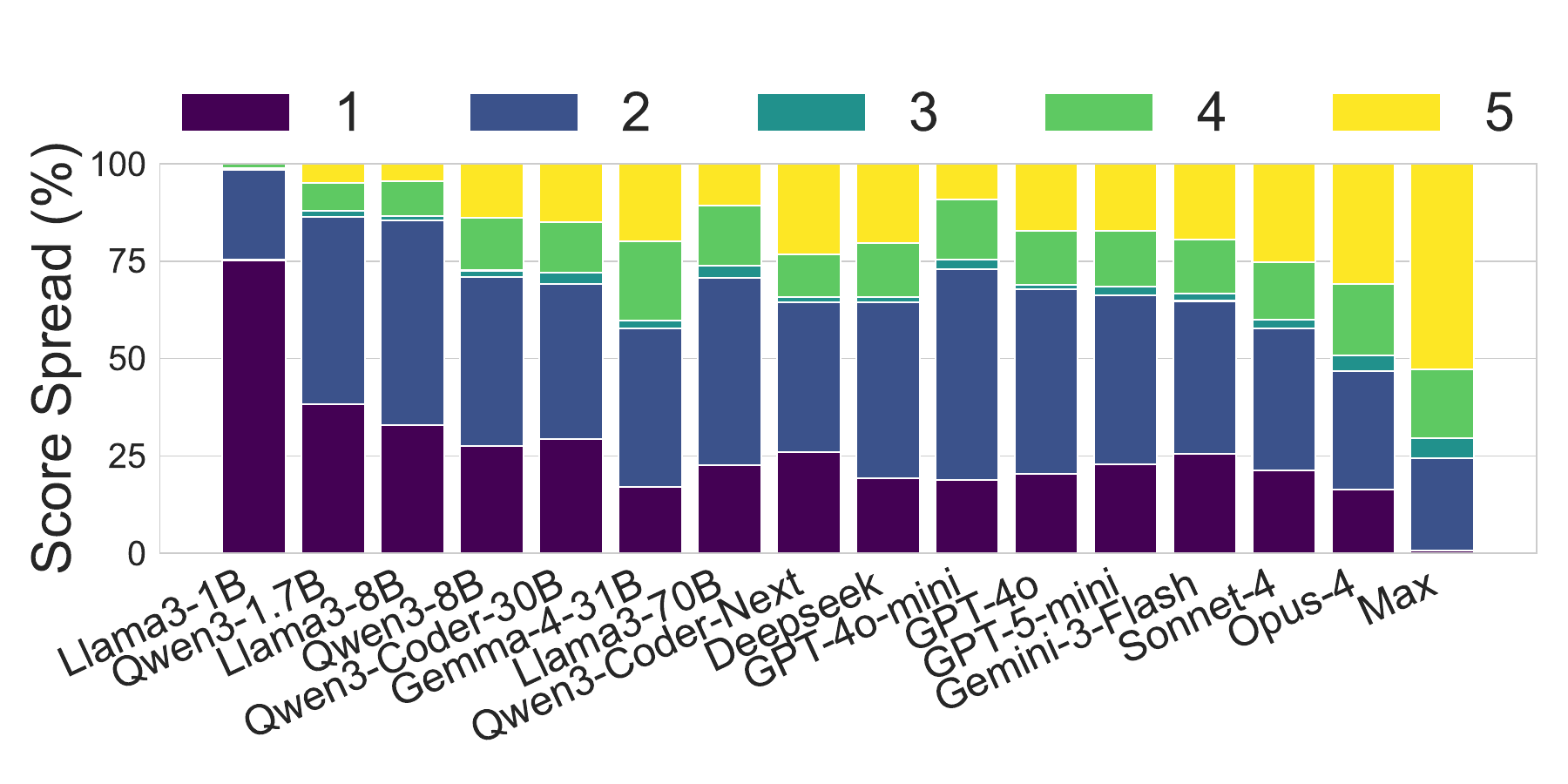}
    \includegraphics[width=0.49\linewidth]{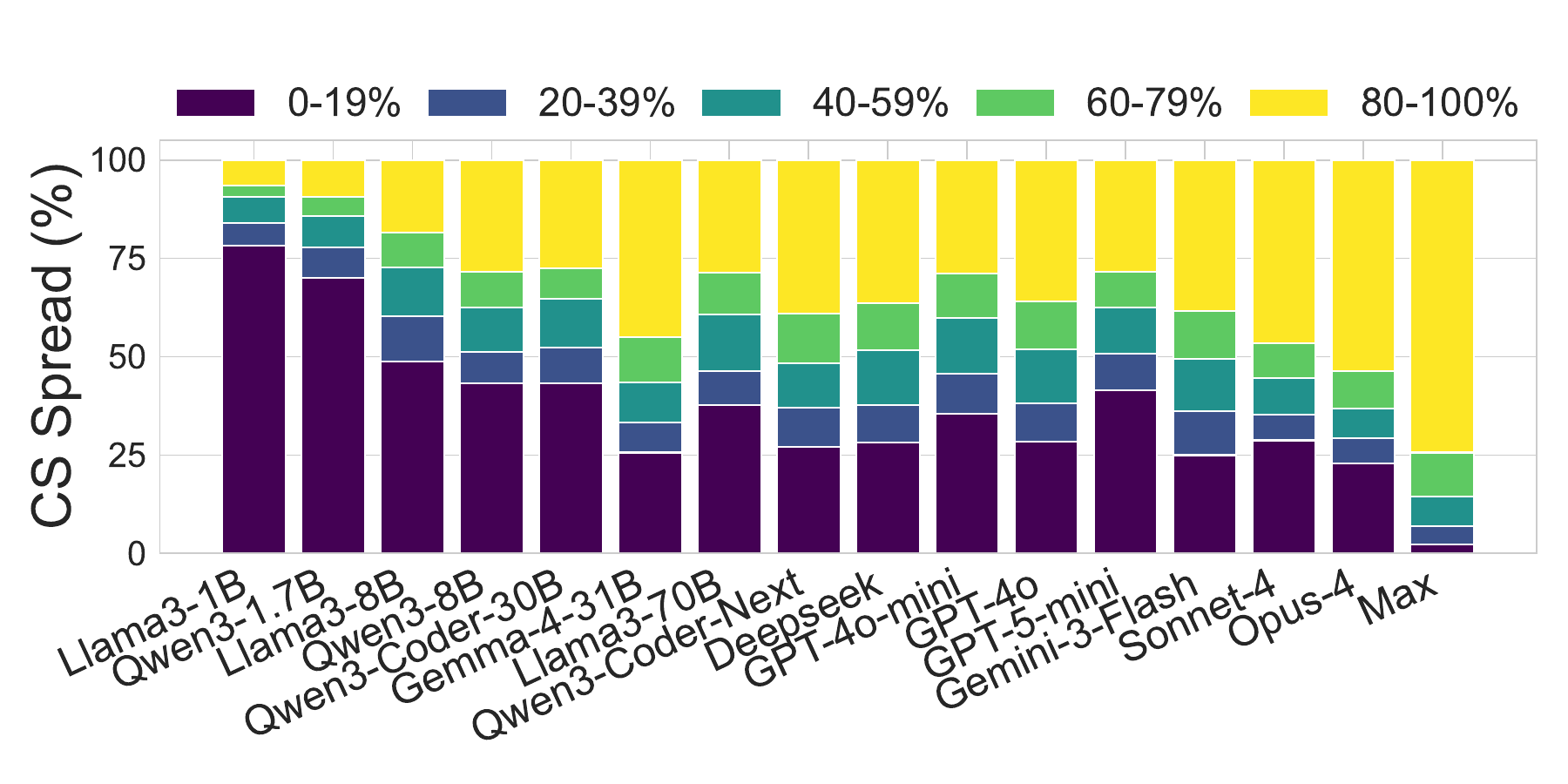}    
    
    \caption{Model performance on \benchmarkshort{} in the interactive setting, with the spread of LM-as-a-judge scores (left) and code match scores (right). Each coloured bar segment represents the \% of the test set for which the description score / code match score falls within that particular range. Even the best performing model (Claude-4-Opus) struggles (score 2 or lower) to accurately describe over 45\% of the methods in the test set. Code written using the predicted description fails (code match score less than 80\%) to completely match the true methods' behaviour on over 45\% of the methods. The `Max' aggregation uses the highest available score (across all models) for a given test instance, establishing an upper bound on current model capabilities and demonstrating the difficulty of \benchmarkshort{}.}
    \label{fig:interactive}
\end{figure}

\subsection{Results}

Regardless of the choice of metric, the performance of the models varies predictably with scale (Figure~\ref{fig:interactive}). Smaller models fail catastrophically while the most powerful, frontier models achieve moderate success (score 4 or above, code match score 60\% or above) on nearly half of the test set. Encouragingly, the rankings implied by the different evaluation methods are highly correlated, with both metrics agreeing on the top models being Claude-4-Opus, followed by Claude-4-Sonnet. This agreement establishes that the multi-aspect evaluation framework identifies clear instances of model improvement. We compute the `Max' performance across all models by selecting the highest score obtained by any model on a given test instance. This post-hoc ensemble still fails to fully understand 25\% of the test set, establishing the difficulty of \benchmarkshort{}. These unsolved instances are not simply impossible: they include cases where the models' final hypotheses are contradicted by their own queries, and which annotators with graduate-level computer science training could reverse engineer through interaction alone (Appendix~\ref{sec:appendix_unsolved}).

We find that different models find different methods challenging to understand. This can be seen in how the `Max' performance is substantially higher than any one individual model. Quantitatively, the correlation between the instance-wise success rate of Gemini-3-Flash and Claude-4-Opus is $0.54$ (Spearman rank correlation~\citep{spearman1961proof}), which falls short of strong agreement. This discrepancy suggests that each model has varying biases that prove helpful or harmful in specific instances. The result also establishes that models can hope to improve substantially on our benchmark. 

However, even Claude-4-Opus struggles (score 2 or lower) to accurately describe over 45\% of the methods in the test set and code written using the predicted description fails (code match score less than 80\%) to completely match the true methods' behaviour on over 45\% of the methods. 

\begin{figure}
    \centering
    \includegraphics[width=0.7\linewidth]{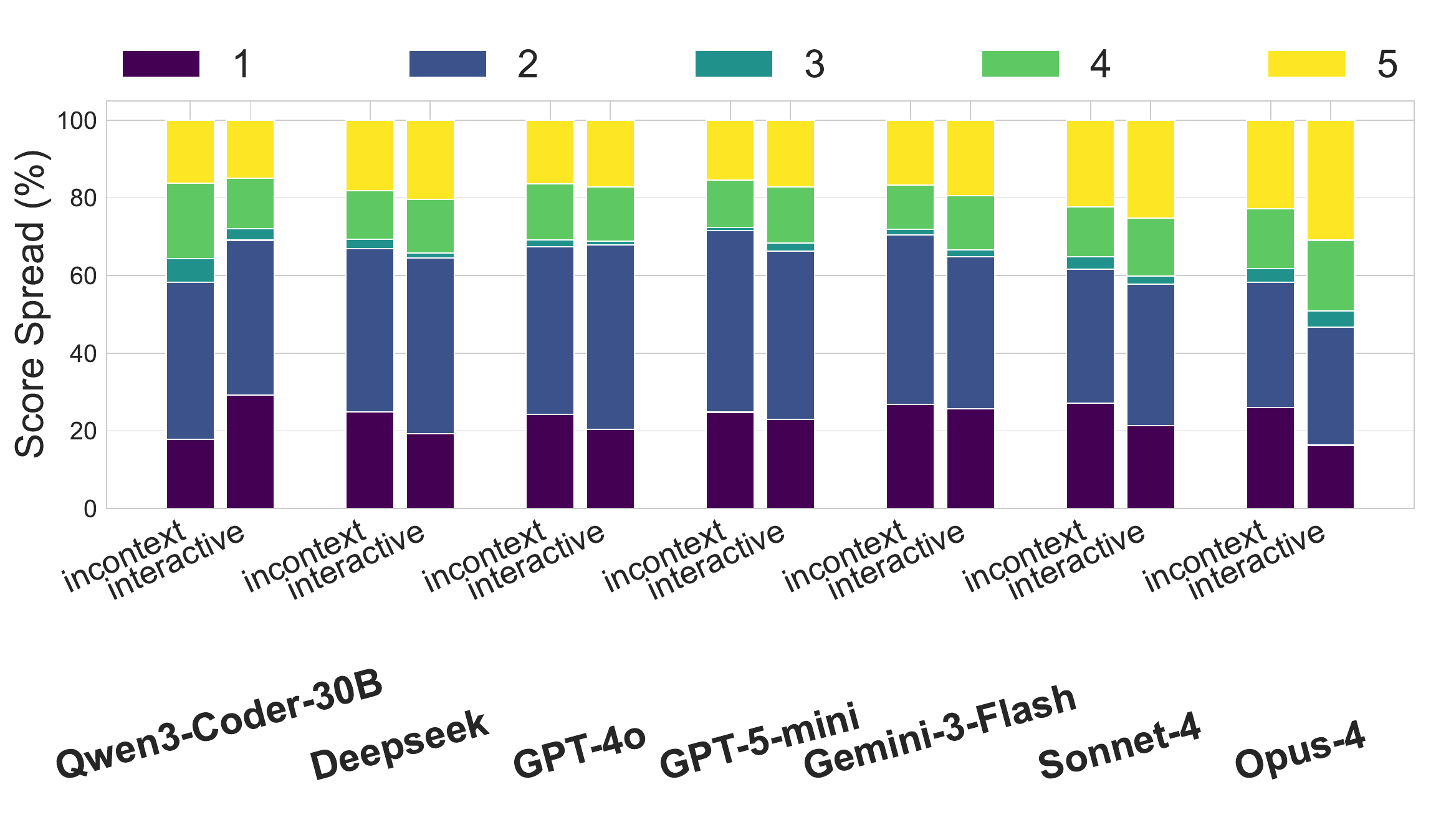}
    \caption{Comparison between the incontext and interactive setting shows that strong models like GPT-4o struggle to fully utilize the benefits of interactive API access. Only the Claude models are significantly (at $p<0.05$) improved in the interactive mode.}
    \label{fig:incontext_partial}
\end{figure}

More concerning is that most models achieve similar performance in the in-context and interactive modes. Using a paired bootstrap t-test~\citep{koehn-2004-statistical}, we find (Figure~\ref{fig:incontext_partial}, Figure~\ref{fig:incontext_results}) no significant difference between the performance of these two settings (at p<$0.05$) for the majority of models evaluated, with the only exceptions being GPT-5-mini, Gemini-3-Flash, DeepSeek, Claude-4-Sonnet and Claude-4-Opus, which significantly improve when given API access. However, even for these models, the degree of difference is marginal. This result also holds for reasoning and code-generation models, which are specifically trained to use tool calls or APIs. The equivalence in performance with and without interaction suggests that most LMs do not effectively use the tool (API) at their disposal to improve their predicted descriptions.  

Overall, our experiments establish that \benchmarkshort{} is challenging for the strongest of models. This poor performance demonstrates that, despite their ability to understand code when provided the text of the function, models struggle to understand code artifacts from their input-output behavior alone. 

\section{Error Analysis}
\label{sec:analysis}

\begin{figure}[h]
    \centering
    \includegraphics[width=0.6\linewidth]{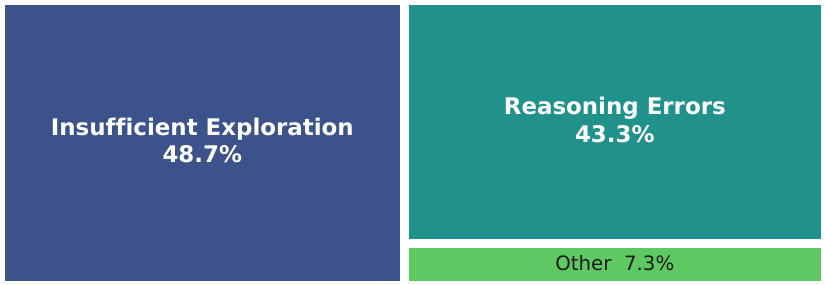}
    \caption{Distribution of failure modes for the three strongest models. Insufficient exploration and reasoning errors account for over 90\% of failures. Format following (0.0\%) and hallucination (0.7\%) errors are virtually absent at this scale and are omitted.}
    \label{fig:failure_modes}
\end{figure}

Having seen that models are unable to efficiently explore the \benchmarkshort{} APIs and determine their true functionality, we perform a qualitative analysis of the successes and failures of various models. Our analysis reveals consistent patterns across models, with specific examples in Appendix~\ref{sec:appendix_interactive}. To quantify these patterns, we use an LM agent (Claude Code~\citep{anthropic2025claudecode}) to classify every failed trajectory of the three strongest models (Claude-4-Sonnet, Claude-4-Opus and Gemini-3-Flash) into one of five categories: format following, hallucination, reasoning errors, insufficient exploration and other (Figure~\ref{fig:failure_modes}).

\noindent\textbf{Format Following}: In order to pass inputs into the API, the LM agent must output a valid Python expression that is evaluated and provided as input arguments to the method. When an LM generates invalid expressions that violate some aspect of Python syntax, we consider this a format following error. While the smallest models in our test suite (Qwen3-1.7B, Llama3-1B, etc) will often fail in this regard, we confirm that the majority of the larger models generate only valid input tuples and do not make format following errors, with none of the failures of the three strongest models attributable to format following.

\noindent\textbf{Hallucination:} In the hypothesis revision stage, the LM must parse the set of input-output examples it has seen, reason over the patterns they imply, and update its running hypothesis. If the LM's reasoning contains references to fabricated input-output examples that it had not seen, we consider it a case of hallucination. Similar to the format following, we observe this trend only in smaller models, with the reasoning chains of more powerful models being consistent with the examples seen and largely free of hallucinations (0.7\% of the failures of the three strongest models).

\noindent\textbf{Reasoning Errors:} In the hypothesis revision stage, the model must determine whether its current hypothesis is satisfactory, and terminate if it is confident. A necessary (but not sufficient) condition for a hypothesis to be sound is that it must fit all of the examples the agent has seen so far. If the agent decides to stop exploring when its current hypothesis does not fit one of the examples it has already seen, we consider it a failure in reasoning and refer to it as overconfidence. We observe consistent overconfidence in small (lower than 4B) and medium-scale models (smaller than 30B), with models often accepting hypotheses that are contradicted by the very examples they have already seen. While this variant of error decreases with scale and reasoning training, it remains pervasive even in the most powerful models, accounting for 43.3\% of the failures of the three strongest models. For example, when exploring a method that returns True iff an uppercase letter is preceded by a lowercase one, Claude-4-Opus observed (`ABC',) $\rightarrow$ False, yet concluded that the method `returns True if the string contains at least one uppercase letter', a hypothesis refuted by an example still visible in its context (Appendix~\ref{sec:appendix_unsolved}). The persistence of this tendency across all models suggests that it is an artifact of more fundamental design decisions in the training of LMs. We posit that this failure occurs due to the tendencies of LMs to lean towards `positive' or `affirmatory' language, i.e., sycophancy~\citep{fanous2025syceval}. Results from verbalised uncertainty quantification also support the claim that LMs are overconfident regarding their decisions~\citep{xiong2024can}, a limitation which may have wide-ranging implications for LM-based agents. 

\noindent\textbf{Insufficient Exploration:} While the above failures occur less as models grow more powerful, we identify a common failure mode that persists with scale, and is the single largest source of failure (48.7\%) for the three strongest models. While exploring APIs, LMs do not actively seek to try inputs that could falsify their current hypothesis or test the edge cases of their hypothesis. This behavioural trait leads to insufficient diversity in attempted inputs, which results in partially correct hypotheses that miss major aspects of an API's functionality. Unlike human software engineers, who often test systems by evaluating edge cases and unexpected inputs ~\citep{godefroid2008automated}, even the most powerful LMs opt for more standard inputs. We consider the selection of strategic exploratory inputs to be the most challenging aspect of the API understanding task, requiring the ability to think creatively about the space of possible methods behind the black-box API~\citep{myers2004art}.   

\section{Asymmetric Actor Critic for Interactive Code API Understanding}
\label{sec:solution}

\begin{figure}
    \centering
    \includegraphics[width=\linewidth]{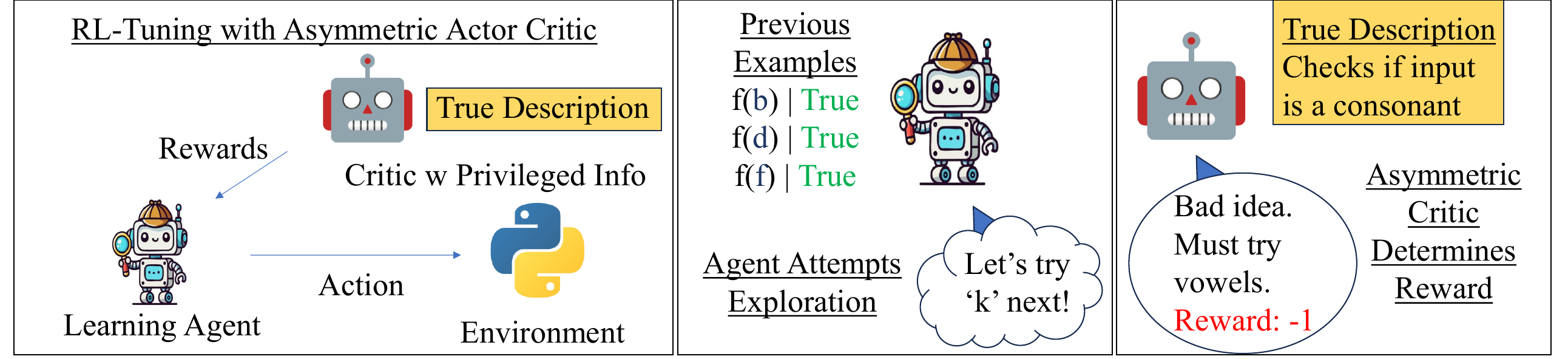}  
    \caption{Overview of RL-Tuning with Asymmetric Actor Critic. At each step of training, the agent's actions are rewarded by a critic that is aware of the gold description which the agent is trying to infer.}
    \label{fig:aac}
\end{figure}
Seeing that models lack the capability to properly explore black-box code APIs, we search for a training paradigm that can imbue them with this ability. The \benchmarkshort{} training set, which matches the structure of the test set (but uses different Python methods), could help facilitate this learning. 

We first evaluate whether the standard paradigm of supervised fine-tuning can help improve models on this task. Specifically, we fine-tune base models (Qwen3-1.7B, Qwen3-8B, Llama3-8B) on the \benchmarkshort{} training set, tuning them to perform description prediction in the \textbf{in-context} mode, i.e. without interactive API access. While we should not expect substantial improvements (consider the example in Figure~\ref{fig:pau}, which can never be reliably resolved without additional interaction), such a fine-tuning process may improve the first guess of the model (based on the two provided examples) and could lead to a minor increase in performance. Unfortunately, we observe no such improvement, with all fine-tuned models performing worse than the base models (in the \textbf{incontext} setting). We identify the core hurdle as the lack of dense, step-wise supervision. An ideal dataset for supervised finetuning shows the model examples of useful, exploratory inputs being attempted and the ideal inferences to draw from the results. However, such a dataset does not exist, and would be prohibitively expensive to collect from human annotation~\citep{ashok-may-2025-little}. More recent RL paradigms, such as Reinforcement Learning with Verifiable Rewards~\citep{DeepSeekAI2025DeepSeekR1IR}, are also challenging to apply in this setting. Unlike an agent writing code (which can be verified with external unit tests), there is no verifiable way to evaluate whether a particular input was `good' for the purpose of exploring the API's functionality. 

For a solution, we take inspiration from works in robot learning, which pioneer the paradigm of Asymmetric Actor Critic methods for Reinforcement Learning~\citep{pinto2017asymmetric, Baisero2021UnbiasedAR}. Traditionally, these methods involve training an agent to optimize rewards as determined by a critic model. The critic model is not necessarily more powerful than the agent (in terms of computational capacity), but rather uses privileged information that the agent does not have to produce enhanced reward signals~\citep{cai2024provable,hu2024privileged}. For example, ~\citet{pinto2017asymmetric} train an agent for pick-and-place manipulation from RGB observations. During training, the critic is given access to full-simulator information such as the ground truth coordinates of the target object, enabling more effective learning. Such a paradigm enables RL algorithms to use information that may be available only during training to enhance the reward signal conveyed to the agent. We adapt this paradigm to our task, and, in doing so, introduce the first successful instantiation of the AAC paradigm for RL-tuning of LMs. 

We create environments (Figure~\ref{fig:aac}) where the agent attempts to explore method APIs from the \benchmarkshort{} train set in the \textbf{interactive} setting. At each step, the agent's actions are rewarded or penalised by a critic that is aware of the gold description which the agent is trying to infer. The critic is frozen over the course of training and uses the same base LM as the agent, obviating concerns that performance improvements are a result of distillation from a more powerful critic model. We design a reward schedule which combines parsing-based reward signals and critic-based rewards to incentivise both format following and deeper exploration. Specifically, in the `input conceptualization' step, we penalize overly long generations and reward the agent if the conceived input is judged to be an effective test of the current hypothesis. For the `input generation' step, we penalize the agent if it generates an input that is not a valid Python tuple or causes an error when used to query the method. Finally, for the `hypothesis revision' step, the rewards depend on the agent's decision to terminate or continue exploration. If the agent's current hypothesis is judged to be accurate, then the agent is penalized for continuing exploration, and if the agent's current hypothesis is judged as inaccurate, the agent is rewarded for continuing exploration. Given these environments, we train the base agent with GRPO~\citep{zheng2023secrets} for a maximum of five epochs, selecting the best checkpoint as judged by average validation-set reward. For exact reward scales, hyperparameters and details, see Appendix~\ref{sec:appendix_aac}.

\subsection{Results}

Across all base models (Figure~\ref{fig:aac}, left, middle), RL-tuning under the AAC paradigm increases the average score ($+0.13$) and average code match score ($+6.27\%$). The AAC-tuned Qwen3-8B model achieves a description score of $2.591$, rivalling the $2.595$ achieved by GPT-5-mini (all results are significant at p<0.05, with a paired bootstrapped t-test). When we use the RL-tuned models in the \textbf{in-context} mode, performance is comparable to the base models. This discrepancy implies that the performance increase from RL-tuning does not come from the model becoming better at making initial guesses about the API's functionality, but rather from more effective exploration and reasoning about outputs. We also investigate the qualitative effects of AAC-tuning on API exploration: 

\begin{figure}[h]
    \centering
    \includegraphics[width=0.32\linewidth]{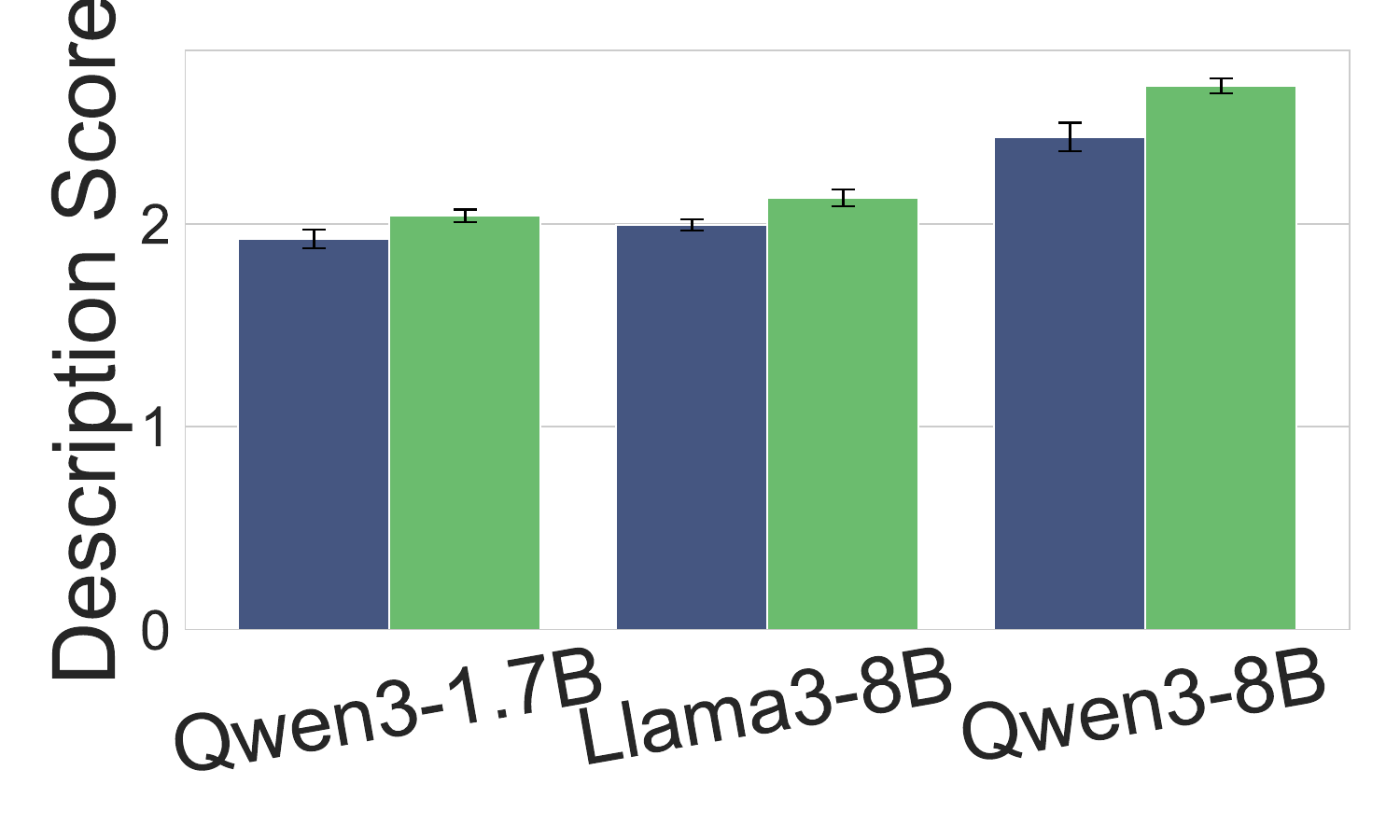}
    \includegraphics[width=0.32\linewidth]{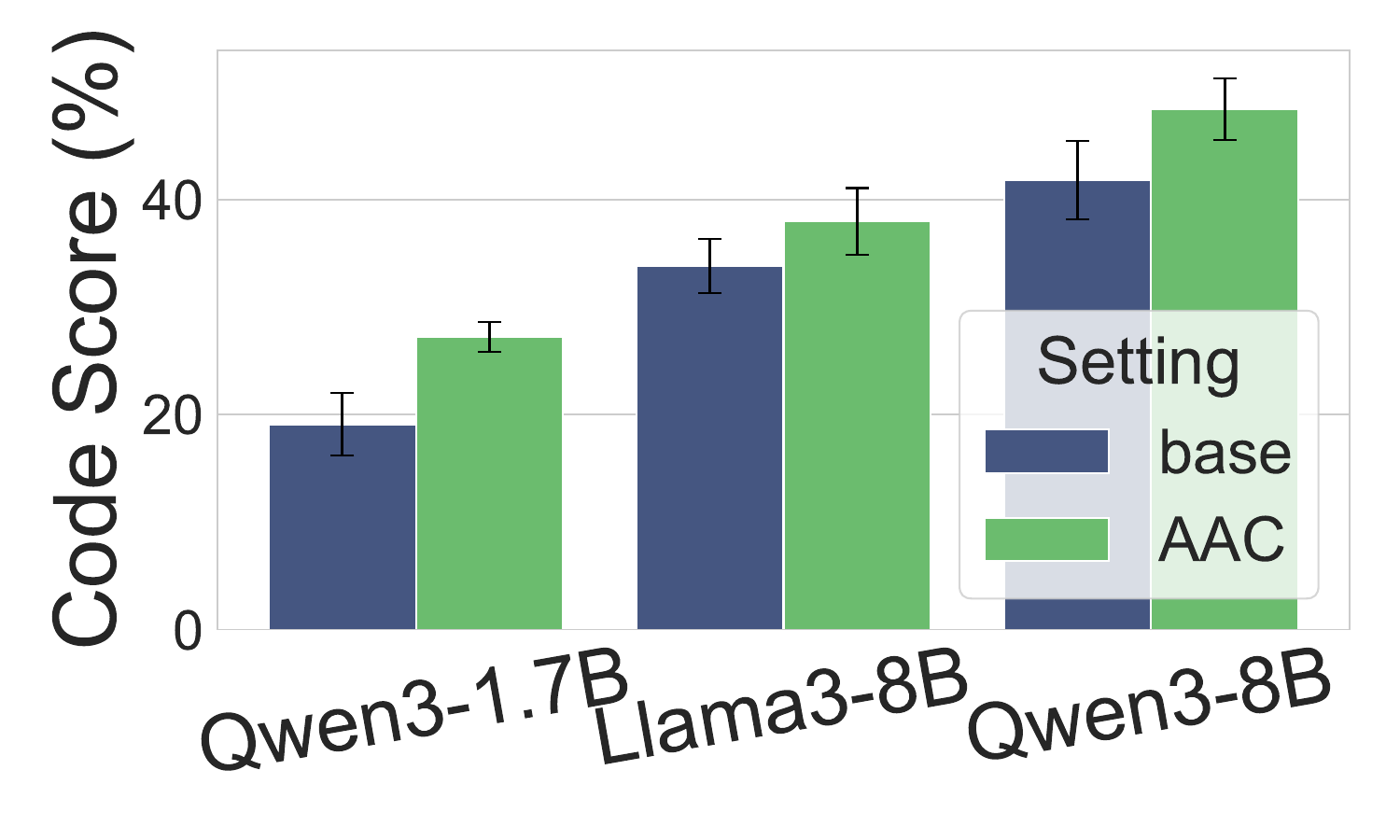}    
    \includegraphics[width=0.32\linewidth]{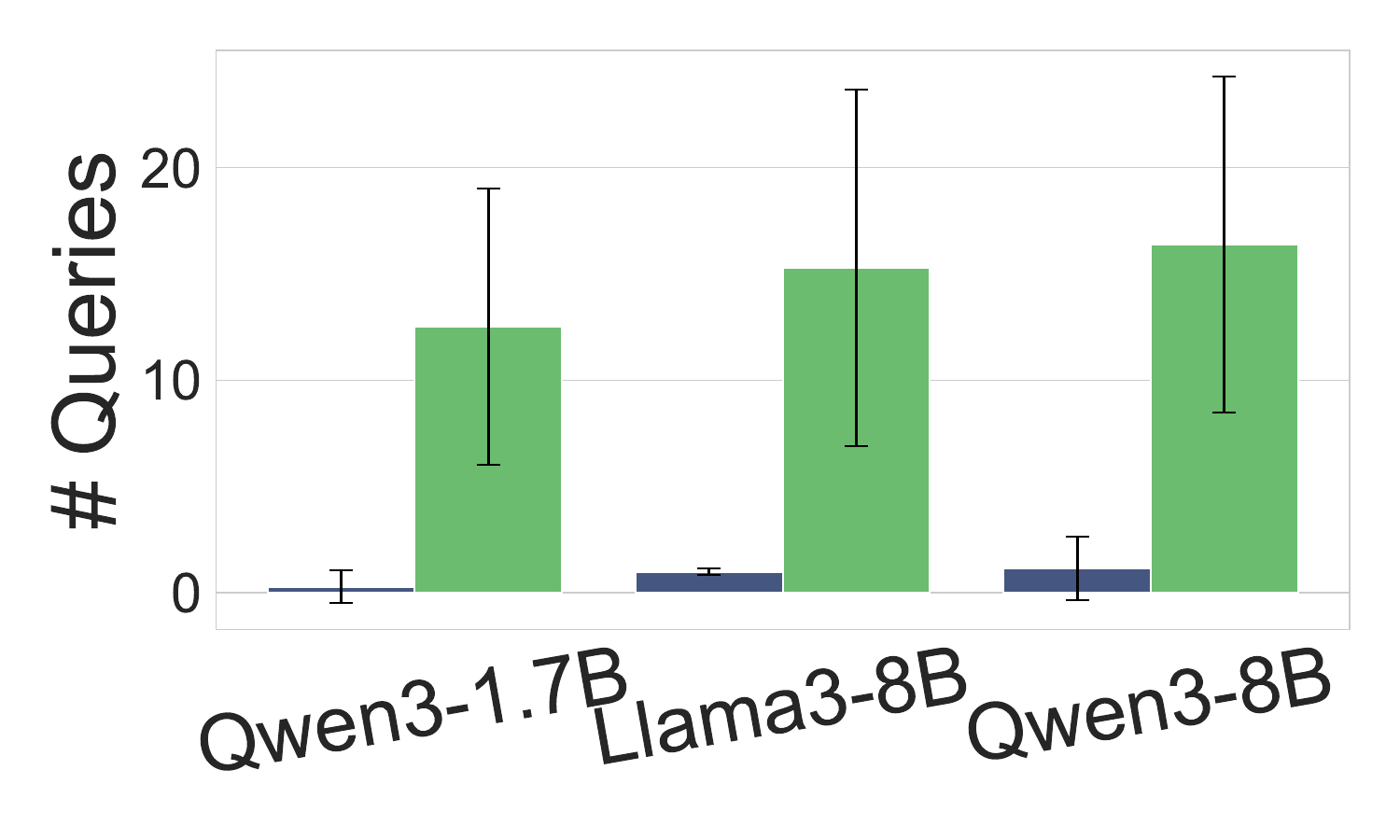}        
    \caption{RL-tuning with AAC makes models better able to explore APIs and understand their functionality (left, middle), significant at $p<0.05$ with a bootstrap t-test. The AAC-tuned Qwen3-8B model achieves a description score of $2.591$, rivalling the $2.595$ achieved by GPT-5-mini. AAC-tuned models make fewer format-following errors and more actively explore the API (right). }
    \label{fig:rl}
\end{figure}

\noindent\textbf{Format Following:} Smaller base models (e.g. Qwen3-1.7B) struggle with following the formatting instructions of the exploration steps and suggest invalid Python tuple expressions as inputs to the API. Due to the environmental reward provided in our training, the RL-tuned variants do not exhibit this behavior and are nearly flawless at generating valid Python expressions with which to query the API. 

\noindent\textbf{Deeper Exploration:} Base models often tend to be overconfident in their current hypothesis and terminate their exploration prematurely. As per our reward schedule, an agent which decides to prematurely terminate is penalised by the critic. This incentivises deeper exploration, and as a result, the RL-tuned models learn to query the API more often, with diverse inputs (Figure~\ref{fig:aac}, right).

We highlight the generality of the Asymmetric Actor Critic paradigm. Several unverifiable tasks, such as creative essay writing, automated data analysis, and strategic decision making, are ones where the final solution to a problem is known; however, the reasoning steps between the problem and solution are hard to annotate or create precise reward signals for. The success of AAC on our task suggests that the method holds promise in such unverifiable domains by leveraging a critic with knowledge of the final solution to provide useful step-level reward signals for optimization.

\section{Related Work}
\label{sec:related-work}
With LMs becoming increasingly proficient at tool-use~\citep{NEURIPS2023_8fd1a81c, patil2024gorilla, huang-etal-2024-planning}, recent benchmarks measure their ability to effectively use growing and diverse libraries of tools~\citep{elder-etal-2026-live}. While existing work measures the abilities of models to effectively use tools~\citep{guo2024stabletoolbench, Dong2025ToolPA} in dynamic contexts~\citep{yu2026benchmarking, trivedi-etal-2024-appworld}, there is limited work on understanding how tools or APIs function through interaction alone. Traditional approaches to black box understanding of APIs come from literature on software testing~\citep{peled2001black}. Classic methods require specification of input and output constraints that an API must follow~\citep{myers2004art}. While such methods can provide powerful formal guarantees~\citep{woodcock2009formal}, they are inapplicable to settings where the APIs do not come with sufficient specifications~\citep{neider2024formal}. Follow-up works seek to learn such specifications through interaction~\citep{Sankaranarayanan2008MiningLS}, however, even in these cases, the exploratory inputs used to discover these specifications are generated from API-specific interface descriptions. Other approaches instead enumerate a variety of possible specifications and perform motivated testing to check whether specific invariants hold over various inputs to the API~\citep{alonso2023agora}. Concurrent work has begun exploring the feasibility of applying LMs to black box API discovery, for example, ~\citet{yang2026apisensor} use LMs and graph algorithms to discover details about a web API from its runtime traffic logs. 

Most similar to our work are \citet{genglarge} and \citet{wei2025codearc}, which evaluate the ability of LMs to reverse engineer Python methods from interaction with a method API or input-output examples. Prior work either adopts a fundamentally different paradigm that does not isolate the failures that \benchmarkshort{} does, or adopts a similar framework but tests on functions that are already discoverable by existing systems.

\citet{wei2025codearc} follow prior work on active learning~\citep{Angluin1987LearningRS} and use an oracle that, with knowledge of the true function, provides counterexamples to each proposed hypothesis and decides when to terminate. This exports the two dominant failure modes we identify (Section~\ref{sec:analysis}), devising exploratory inputs and judging when to stop, to the oracle, and is an unrealistic model of APIs in the wild. Fine-tuning on their answer-aware trajectories is also harmful in our setting (Appendix~\ref{sec:appendix_aac_baselines}).

The setup of \citet{genglarge} is similar to \benchmarkshort{}, but their test set is limited to 100 list-mapping programs that implement a lambda expression, where the query is a list of integers and the response is an integer. Compared to \benchmarkshort{}, which involves a varying number of input arguments, diverse input and output types, and more complex internal logic, this test set lacks scale and diversity of inputs, outputs and functionality. More importantly, it is no longer challenging for frontier LMs: when we run the 80 unique programs from \citet{genglarge} in the \benchmarkshort{} interactive setting, Gemini-3-Pro~\citep{team2024gemini} achieves a description score of 4.13/5, with a perfect score on 87.5\% of the programs. Neither of these works proposes a solution to the problems they set up, while we show that our novel adaptation of the AAC paradigm can improve the ability of models to perform interactive API understanding.

\section{Limitations and Ethical Considerations}
\label{sec:limitations_future}
As a benchmark for Python API methods, our work does not evaluate the entire range of possible black-box APIs, with important code-related functionality (such as memory management) not evaluated due to its reduced importance in Python. We additionally only use stateless methods in our benchmark, a simplification of the interactive code understanding problem. Despite these decisions, our evaluations demonstrate the difficulty of \benchmarkshort{}, suggesting that even simple instantiations of interactive code understanding are challenging for frontier LMs. Similarly, since the test methods are sourced from popular code benchmarks, they may have been exposed during LM pre-training (see Appendix~\ref{sec:contamination}). We believe that these factors only accentuate the surprise of our findings: even stateless, potentially pre-exposed functions behind simple APIs cannot be reliably understood through black-box interaction. We view solving \benchmarkshort{} as a necessary, but not sufficient, prerequisite for agents that can understand code through black-box interaction alone. Just as simple code synthesis benchmarks like HumanEval~\citep{chen2021evaluating} and MBPP~\citep{austin2021program} helped build the scaffold for systems that eventually perform well on sophisticated testbeds like SWE-Bench~\citep{jimenez2024swebench}, we hope that \benchmarkshort{} will serve as a fundamental testing ground for black-box understanding of code, enabling rapid iteration on methods that will prove useful for more complex and realistic settings.

While interactive code understanding is a capability that underlies fundamental workflows in software engineering~\citep{kery2017exploring}, the applications of this skill include reverse engineering of proprietary APIs~\citep{shatnawi2017reverse}. Our results establish that reverse engineering of sophisticated APIs through interaction alone is likely out of the capabilities of existing models. However, a stronger generation of models may enable actors to illicitly extract private code, and must be considered when developing models capable of deep exploration and interactive understanding.

\section{Conclusion}
\label{sec:conclusion}

In this work, we introduce the \benchmarkshort{} benchmark, a testbed for interactive understanding of code APIs. We demonstrate that despite the impressive performance of coding agents on standard code understanding benchmarks, they consistently fail on \benchmarkshort{}, with no model able to understand more than 45\% of the test APIs. Our analysis reveals the consistent failure modes of overconfidence and premature termination of exploration, a problem that persists across various LMs with different training paradigms, suggesting a more fundamental limitation of LM-based agents. Finally, we adapt the Asymmetric Actor Critic paradigm from robot learning to RL-tuning of LMs for interactive code understanding, leading to models that conduct more active exploration of APIs and prove more capable at understanding their functionality.

\bibliographystyle{abbrvnat}

\newpage

\bibliography{main}

\newpage

\appendix

\section{\benchmark{} Benchmark}
\label{sec:appendix_benchmark}

\subsection{Source datasets}
\benchmarkshort{} consists of 3.9K train and 741 test instances, with their function bodies sourced from pre-existing benchmarks:

\noindent\textbf{HumanEval} consists of 164 hand-written Python programming problems released by OpenAI~\citep{chen2021evaluating}. Each task includes a function signature, docstring, body, and unit tests, specifically designed to evaluate a model's ability to solve self-contained algorithmic challenges. It is released under the MIT License.

\noindent\textbf{MBPP} (Mostly Basic Python Problems) contains 974 crowd-sourced tasks curated by Google Research~\citep{austin2021program}. These problems are designed for entry-level programmers and include a natural language description, a reference solution, and three automated test cases to verify functional correctness. It is released under the CC-BY 4.0 license.

\noindent\textbf{CruxEval} is a benchmark of 800 Python functions paired with input-output sets, developed to assess code reasoning and execution capabilities~\citep{gu2024cruxeval}. It requires models to perform both "input prediction" and "output prediction" tasks, testing understanding of program behavior beyond simple generation. It is released under the MIT License.

\noindent\textbf{CodeAlpaca} comprises 20,000 instruction-following samples generated using the self-instruct method for code-related tasks~\citep{codealpaca}. It covers a diverse range of programming activities, including code generation, editing, and optimization, to facilitate instruction tuning in large language models. It is released under the Apache License 2.0. This dataset is used to generate the training set alone. 

\subsection{Data Cleaning}
\noindent\textbf{Function Identification} Raw functions are extracted from each source dataset and filtered to retain only well-formed Python functions. For datasets that may contain multi-function files (e.g., \textsc{HumanEval}, \textsc{MBPP}), we isolate the primary target function while preserving any necessary helper definitions as preamble code. Functions involving non-Python constructs (e.g., Ruby \texttt{end} keywords or C-style braces) and functions that do not begin with a \texttt{def} statement are discarded at this stage.

\noindent\textbf{Function Anonymization:} To prevent models from identifying functions by name or signature rather than behavior, each function undergoes a standardized anonymization procedure.
First, all \texttt{import} statements are hoisted to the top of the function body.
Docstrings are then stripped to remove any natural-language descriptions of the function's behavior.
Functions that import unsafe system-level modules (\texttt{os}, \texttt{sys}, \texttt{subprocess}, \texttt{shutil}, \texttt{socket}, \texttt{requests}) are discarded to avoid side effects during execution.
The function name is renamed to \texttt{test\_func}, and all argument names are replaced with positional aliases \texttt{arg0}, \texttt{arg1}, \ldots, \texttt{arg$N$}.
Return type annotations are stripped from the header.
To preserve readability of the body under the original argument names, each anonymized argument is re-bound at the start of the function body via assignments of the form \texttt{original\_name = arg$i$}.
Finally, a call to a stub \texttt{validate\_input\_args} function is prepended to the body; this stub is later filled in by the validation stage.

\begin{figure}
\begin{tcolorbox}
You are given a function definition with several arguments. 
Your task is to first, identify the types and other fundamental 
constraints of the input variables that are required for the 
function to run without errors. Then, create a validate\_input\_args 
function that checks these constraints and raises appropriate 
exceptions if any of them are violated.
Function: 
\begin{lstlisting}
test_func(arg0: List[float], arg1: float) -> bool:
    threshold = arg1
    numbers = arg0


    for idx, elem in enumerate(numbers):
        for idx2, elem2 in enumerate(numbers):
            if idx != idx2:
                distance = abs(elem - elem2)
                if distance < threshold:
                    return True

    return False
\end{lstlisting}
Validation Function: 
\begin{lstlisting}
def validate_input_args(arg0: List[float], arg1: float):
  if not isinstance(arg0, list):
        raise TypeError("arg0 must be a list")
  for item in arg0:
    if not isinstance(item, float):
        raise TypeError("All elements in arg0 must be floats")
    if not isinstance(arg1, float):
        raise TypeError("arg1 must be a float")
    return     
\end{lstlisting}
[STOP]
Test Function: 
\begin{lstlisting}
def test_func(arg0, arg1):
    \"\"\"
    Find the similar elements from the given two tuple lists.
    \"\"\"
    res = tuple(set(arg0) & set(arg1))
    return (res)
\end{lstlisting}    
Validation Function:
\begin{lstlisting}
def validate_input_args(arg0: tuple, arg1: tuple) -> None:
    if not isinstance(arg0, tuple):
        raise TypeError("arg0 must be a tuple")
    if not isinstance(arg1, tuple):
        raise TypeError("arg1 must be a tuple")
    return     
\end{lstlisting}
[STOP]
Now, generate the validate\_input\_args function for the following 
function only, from the def validate\_input\_args portion to the 
return line. Make sure to include type annotations on the function 
definition. After that, say [STOP]
Test Function: 

\caption{Validation function creation prompts}
\label{fig:creation_prompts}
\end{tcolorbox}
\end{figure}

\begin{figure}
\begin{tcolorbox}
You are given a function definition. Your task is to create as many 
example inputs as you can to the function that satisfy the 
constraints, and also trigger the different branches of the function 
logic. Output as many examples as you can, that test different 
parts of the function.     
Output each example on a new line, in the format:
Reasoning: An extremely brief reasoning for the kind of behavior 
these examples will trigger
 - (arg0, arg1, ..., argN)
 - (arg0, arg1, ..., argN)
Reasoning: brief reasoning for kind of behavior
 - (arg0, arg1, ..., argN)
 - (arg0, arg1, ..., argN)
 - (arg0, arg1, ..., argN)
 - (arg0, arg1, ..., argN) 
[STOP]
Function:
\begin{lstlisting}
def validate\_input\_args(arg0):
    if not isinstance(arg0, int):
        raise TypeError("arg0 must be an integer")

def test\_func(arg0):
    """
    Identify non-prime numbers.
    """
    validate\_input\_args(arg0)
    result = False
    for i in range(2,int(math.sqrt(arg0)) + 1):
        if arg0 % i == 0:
            result = True
    return result
\end{lstlisting}
Reasoning: Since the function tests prime numbers, and takes 
arguments a single integer, we first test with small prime numbers
 - (2)
 - (3)
 - (17)
 - (19)
Reasoning: We can also test with small non-prime numbers
 - (4)
 - (6)
 - (21)
 [STOP]

Note, the type checks in validate\_input\_args are bindings. 
So you must always ensure that the inputs you are generating satisfy 
those type checks. For example, if validate\_input\_args checks that 
an argument is a float, you MUST give a float in your examples, not an int. 
Now do this for the following function only. After that, say [STOP].
Function: 
\end{tcolorbox}
\caption{Example creation prompts}
\label{fig:example_prompts}
\end{figure}

\noindent\textbf{Validation Function Generation and Executability Filtering:}
A two-phase filtering step ensures that each anonymized function can be executed deterministically.
In the first phase, a language model generates a \texttt{validate\_input\_args} function for each anonymized function.
The model is prompted with the original (pre-anonymization) function body as context, and is asked to identify the types and structural constraints on each argument and emit appropriate \texttt{isinstance} checks with typed exceptions.
The generated validator is prepended to the anonymized function, forming the complete \texttt{test\_func\_validated} artifact.
Any function for which the combined code fails to compile or instantiate as a callable is dropped.

In the second phase, we separately verify that each function is executable in isolation: we attempt to load each function into a sandboxed runner without any input, and discard any function whose definition itself raises an exception at load time.
Functions that create new files or directories as a side effect of initialization are also detected and removed by comparing the working directory listing before and after instantiation.

\noindent\textbf{Example Generation and Train/Test Split:} For each surviving function, a language model is prompted to generate diverse input tuples that exercise different branches of the function logic.
Generated inputs are filtered to retain only those that (i) satisfy the \texttt{validate\_input\_args} constraints and (ii) execute without error, yielding a valid, serializable output.
Functions for which no valid input can be produced are discarded.

Retained examples are then partitioned into train and test splits.
We allocate $n_\text{train} = 2$ examples as few-shot demonstrations and require at least $n_\text{test} = 5$ held-out test examples per function; functions with insufficient coverage are dropped. To promote output diversity within the training set, we enforce that no two training examples share the same output value, and that at most one training example exhibits an identity mapping (i.e., input equals output).
A final round-trip JSON serialisation check removes any rows that cannot be losslessly serialised and deserialised, ensuring downstream compatibility.

All Language Model calls in the above pipeline use GPT-4o. The prompts for these steps are given in Figure~\ref{fig:creation_prompts} and Figure~\ref{fig:example_prompts}.

\noindent\textbf{Coverage-based Filtering}
To verify that the generated test examples exercise the function's logic rather than a single execution path, we measure branch coverage using Python's \texttt{coverage} library.
For each function, we write a self-contained script that imports the function, iterates over all held-out test examples, and calls \texttt{test\_func} on each input.
We then run coverage analysis across all functions, combine the parallel traces, and report per-branch coverage.
Functions whose test examples achieve less than full branch coverage are flagged; the pipeline achieves 99\% branch coverage on average across the retained benchmark.

\subsection{Final Benchmark}

\begin{figure}
    \centering
    \includegraphics[width=0.75\linewidth]{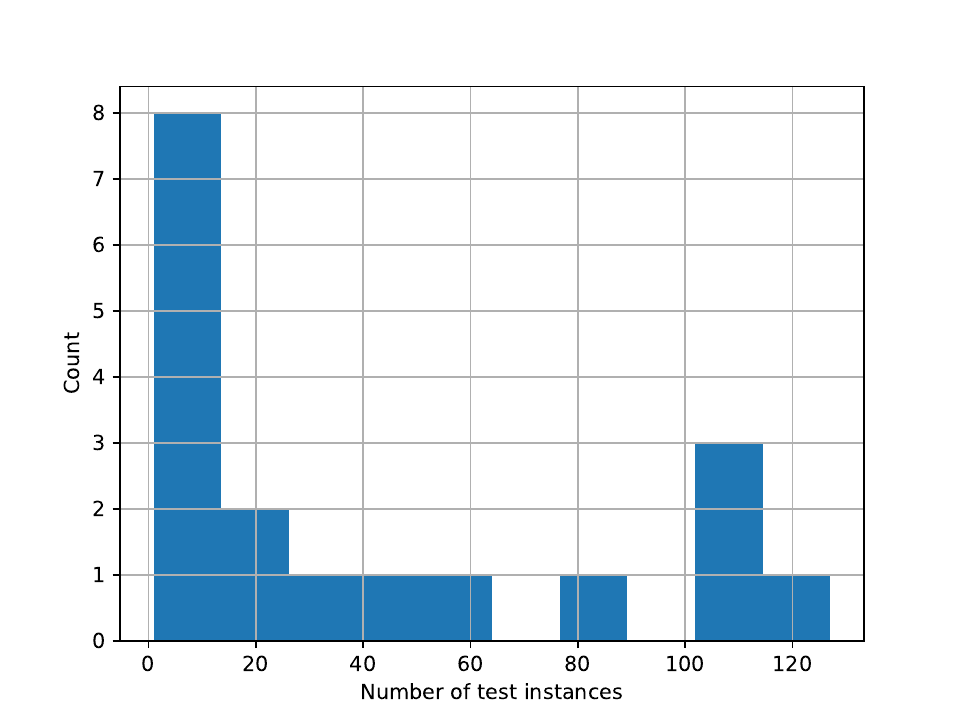}
    \caption{Histogram over test suite sizes}
    \label{fig:test_suite_size}
\end{figure}

After the data cleaning, we are left with 741 instances of Python methods that can be queried with inputs and either return a value or raise TypeError with respect to the input parameters. Each contextual set has exactly two examples, which trigger different outputs from the method. The number of examples in the test suite varies by method, with an average test suite size of $7.45$. We show examples in Figure~\ref{fig:bench_examples} and provide a histogram over test suite sizes in Figure~\ref{fig:test_suite_size}.

\begin{figure}[th]
\begin{tcolorbox}
\textcolor{blue}{Code:}
\begin{lstlisting}
def test_func(arg0):
    validate_input_args(arg0) # raise int TypeError
    n = arg0
    if n%2 == 0:
        return True
    else:
        return False
\end{lstlisting}

\textcolor{blue}{Description:}
Checks if the given number 'arg0' is even. It returns True if the number is even and False if it is odd.

\textcolor{blue}{Contextualizing Examples:}
[["(4)", true], ["(-7)", "False"]]

\textcolor{blue}{Test Suite:}
	
[["(-2)", true], ["(-3)", false], ["(7)", false], ["(15)", false], ["(3)", false], ["(2)", true], ["(-4)", true], ["(10)", true], ["(16)", true], ["(1)", false]]

\rule{\linewidth}{0.4pt}
\textcolor{blue}{Code:}
\begin{lstlisting}
def test_func(arg0, arg1):
    validate_input_args(arg0, arg1) # raise str typerror 
    # and positive int value error
    key = arg1
    text = arg0
    
    
    ciphertext = ""
    for char in text:
        if not char.isalpha():
            ciphertext += char
            continue
        offset = ord('a') if char.islower() else ord('A')
        new_char = (ord(char) - offset + key) % 26
        ciphertext += chr(offset + new_char)
return ciphertext
\end{lstlisting}

\textcolor{blue}{Description:}
Ciphers a given text using the Caesar cipher technique, where each letter in the input text 'arg0' is shifted by a specified integer 'arg1'. Non-alphabetic characters are preserved unchanged. The function returns the resulting ciphertext.

\textcolor{blue}{Contextualizing Examples:}
[["(\"hello\", 3)", "khoor"], ["(\"aBcDeF\", 1)", "'bCdEfG'"]]

\textcolor{blue}{Test Suite:}
[["(\"Test\", 0)", "Test"], ["(\"123 and ABC\", 4)", "123 erh EFG"], ["(\"XyZ\", 2)", "ZaB"], ["(\"World\", 5)", "Btwqi"], ["(\"Hello, World!\", 2)", "Jgnnq, Yqtnf!"]]

\end{tcolorbox}
\caption{Examples from \benchmarkshort{}}
\label{fig:bench_examples}
\end{figure}

We present summary statistics regarding the degrees of the methods in our benchmark in Figure~\ref{table:datastat}.

\begin{table}[]
\centering
\caption{Data statistics for counts of methods by number of input arguments in \benchmarkshort{}.}
\label{table:datastat}
\begin{tabular}{@{}lr@{}}
\toprule
No of Input Args & Count \\ \midrule
1        & 377   \\
2        & 311   \\
3        & 52    \\
4        & 1     \\ \bottomrule
\end{tabular}
\end{table}

\subsection{Data Contamination}
\label{sec:contamination}
Owing to the popularity of the benchmarks we use to source the methods in the test set, there is a high possibility that their method descriptions have been leaked in LM training corpora~\citep{xu2024benchmark}. However, we note that since the evaluation setting of \benchmarkshort{} does not provide the model with the original function names or the description, the impact of such contamination is limited. As indicated by Table~\ref{table:datastat}, there are several methods with the same number of input arguments, and hence the header does not provide sufficient information to resolve the method description, even if it was pre-exposed to the Language Model. Additionally, the contextualizing examples are not taken from any benchmarks, but generated for \benchmarkshort{} specifically, implying that the LMs do not have pre-trained associations between the contextualizing examples and the true description of the methods. Ultimately, the poor performance of frontier models on \benchmarkshort{} gives the most substantial evidence that even if the predictions were pre-exposed during LM training, the \benchmarkshort{} benchmark remains meaningfully challenging. For a discussion of how potential contamination and the simplicity of the test methods relate to the scope of our findings, see Section~\ref{sec:limitations_future}.

\subsection{Addressing Impossibility of API Discovery:}
\label{sec:appendix_unsolved}
Since the method arguments and return types can be highly compositional with respect to the basic Python types (e.g. Dictionaries of dictionaries of lists, etc.), type analysis is challenging to perform on aggregate. Over the test bed, the argument and return types of the methods include strings, integers, floats, dictionaries, lists, sets and compositions of these fundamental types. 

In general, the task of understanding a code API purely through black-box interaction is computationally intractable~\citep{peled2001black, alonso2023agora}. Consider the test API:
\begin{lstlisting}
def test(x):
    return x == random.uniform(-1, 1)
\end{lstlisting}

This method is practically impossible to fully characterize, as an agent seeking to describe its behavior with a finite number of queries would (in expectation) never observe an output of `True'. 

However, we provide evidence that the poor performance of models on our benchmark is not due to the impossibility of the task, but rather an inability to effectively explore black box APIs. 

\begin{figure}
    \centering
    \includegraphics[width=0.75\linewidth]{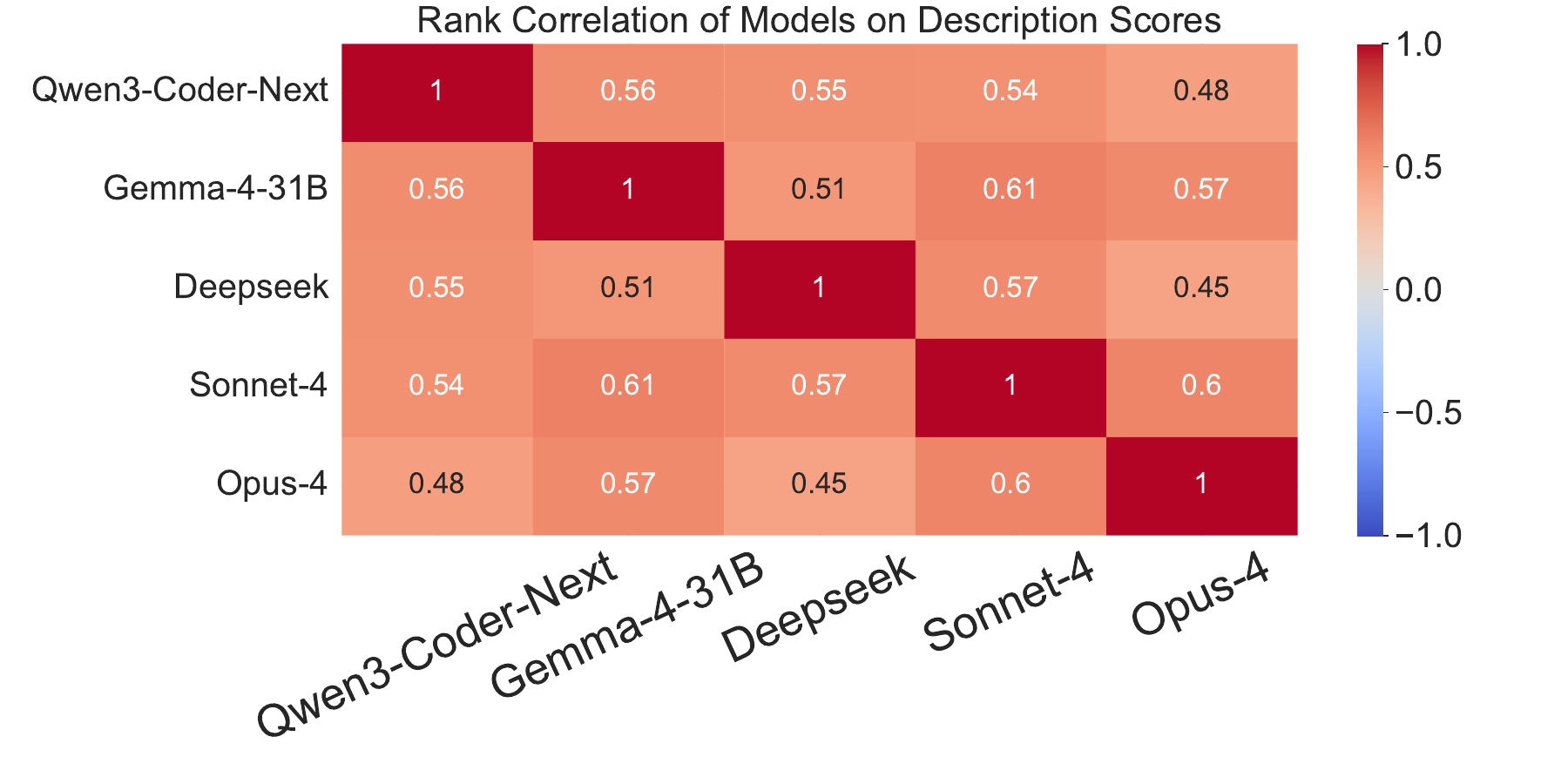}
    \caption{Spearmann Correlation between ranks of test instances across models. Moderate correlation measures suggest that different models are successful at different test instances.}
    \label{fig:correlation}
\end{figure}

The first quantitative proof of this fact can be seen in the performance of the MAX aggregation setting (Figure~\ref{fig:interactive}), which achieves considerably higher scores than the individual model. Another way of viewing this result is through the correlation between the model success rates per test instance (Figure~\ref{fig:correlation}). Both of these results establish that different models struggle with different test instances, suggesting that there is considerable scope for improvement. At the very least, models can reasonably be expected to match the performance of the MAX setting, which would already require a considerable improvement in model capability.  

Our inspection of the instances that remain unsolved by every model also reveals clear cases of reasoning errors and insufficient exploration, suggesting that the flaw lies in the model rather than in the impossibility of the instance (Figure~\ref{fig:unsolved_examples}). Two annotators with graduate-level computer science training were able to reverse engineer the correct description of both instances shown through interaction alone. However, we acknowledge that we cannot guarantee that every test instance is genuinely recoverable through black-box interaction. We can state with confidence that the MAX setting demonstrates substantial headroom for every individual model on \benchmarkshort{}, and that our qualitative analysis of error traces on unsolved instances suggests that the performance of the MAX setting itself can improve with more powerful models.

\begin{figure}
\begin{tcolorbox}
\textbf{Instance 214}

\textcolor{blue}{True Description:} Checks if the input string `arg0' contains at least one uppercase letter that is preceded by a lowercase letter. It returns True if such a condition is met; otherwise, it returns False.

\textbf{Contextual Examples}

("TestString",) => True,

("12345",) => False

\textbf{Explored Inputs}:

("Test123",) => True,
("hello",) => False,
("Hello",) => True,
("a",) => False,
("hEllo",) => True,
\textbf{("ABC",) => False}

\textcolor{red}{Predicted Description}: the function returns `true' if the input string contains at least one uppercase letter, and `false' otherwise. [Incorrect, contradicted by ("ABC",) => False]

\rule{\linewidth}{0.4pt}

\textbf{Instance 283}

\textcolor{blue}{True Description:} Creates and returns a new dictionary that combines a specified number of copies (indicated by `arg1') of the input dictionary (`arg0'), effectively merging them into a single dictionary. If `arg1' is 0, an empty dictionary is returned.

\textbf{Contextual Examples}

(\{'k': 0\}, 10) => \{'k': 0\},

(\{'k1': 100, 'k2': 200\}, 5) => \{'k1': 100, 'k2': 200\}

\textbf{Explored Inputs} (selected, 19 in total):

(\{'a': -1, 'b': 0, 'c': 5\}, 0) => \{\},
(\{'a': 1, 'b': 2, 'c': 3\}, 2) => \{'a': 1, 'b': 2, 'c': 3\},

\textbf{(\{'a': 1, 'b': 2, 'c': 3, 'd': 4, 'e': 5\}, 3) => \{'a': 1, 'b': 2, 'c': 3, 'd': 4, 'e': 5\}},

\textbf{(\{'a': 3, 'b': 15, 'c': 7\}, 1) => \{'a': 3, 'b': 15, 'c': 7\}},
(\{\}, 5) => \{\}

\textcolor{red}{Predicted Description}: the function returns the first `arg1' items from the dictionary. [Incorrect, contradicted by the highlighted inputs]
\end{tcolorbox}
\caption{Instances of \benchmarkshort{} that remain unsolved by every model, with trajectories from Claude-4-Opus. In both cases, the final hypothesis is directly contradicted by examples the model itself queried (bold), which remained visible in its context. Two annotators with graduate-level computer science training were able to recover the correct description of both methods through interaction alone.}
\label{fig:unsolved_examples}
\end{figure}

\section{Evaluation Metrics}
\label{sec:evalrobust}

\begin{figure}[ht]
\begin{tcolorbox}
\textcolor{blue}{LLM-as-a-judge:}
You are given a function description and a hypothesized description of what the function does.
Your task is to rate how accurate the hypothesized description is compared to the true description on a scale from 1 to 5, where 1 means "completely inaccurate" and 5 means "completely accurate".
First, provide an extremely brief explanation (1 sentence) of why you gave that rating. Then, provide your rating in the format "Rating: X" where X is an integer between 1 and 5.
Example:
True Function Description: This function takes a list of integers and returns True if there are any two integers in the list that sum to zero, otherwise it returns False.
Hypothesized Description: This function checks if there are two numbers in the list that add up to zero.
Explanation: The hypothesized description accurately captures the functionality of the true description.
Rating: 5 [STOP]

True Function Description: calulates the nth fibonacci number
Hypothesized Description: This function computes the factorial of a number.
Explanation: The hypothesized description is incorrect as the fibonacci sequence and factorial are different mathematical concepts.
Rating: 1 [STOP]

Now, provide your rating for the following description only. You absolutely must follow the format shown in the examples above and no matter what, you must provide a rating between 1 and 5.
True Function Description: [TRUE]
Hypothesized Description: [HYPOTHESIS]
Explanation (very short):

\textcolor{blue}{Code Writing Model:}
You are an expert programmer. Your goal is to create a Python function called `test\_func` that matches the following description:
[DESCRIPTION]

The header of the function must be:
[HEADER]

The function must satisfy the following input-output examples:
[EXAMPLES]

Now, write the complete code for the function `test\_func` that meets the above requirements. You should structure your response as follows:
Reasoning: <any brief thinking or reasoning you want to do before writing the code. This must be extremely brief and concise, just a sentence or two at most>
Code:
```python
<your code here>
```
[STOP] \# make sure to include the [STOP] token at the end of your response to indicate that you have finished writing the code.
Now, provide your reasoning and code below, and remember to end with [STOP].
Reasoning:

\end{tcolorbox}
\caption{Prompts used by judge and code writing models for evaluation}
\label{fig:eval_prompts}
\end{figure}
We provide the exact prompts used by the LLM-as-a-judge and description-to-code models in Figure~\ref{fig:eval_prompts}.

\subsection{Validating Robustness of Evaluation Metrics}

Given our chosen LLM-as-a-judge model (Llama3-70B) and code-description model (Qwen3-Coder-30B-A3B), we conduct additional checks to verify the robustness of its score ratings. For all experiments below, we use the predicted descriptions for the DeepSeek, Gemini-3-Flash, GPT-4o and Claude-Opus models. 

\noindent\textbf{Robustness to prompt variation:} The official prompt for the score judge provides two specific few shot examples to help guide score judgements. We additionally randomly sample five fresh pairs of (predicted description, true description) tuples from the available predicted descriptions, annotate them with a rating and provide these pairs as fewshot examples to make five alternate prompts. For the code generation model, we write five alternate prompts that request a generated description. We observe that using a different prompt has no effect on the relative ranking of the models by score, with the variance of the average score across different prompts (description score variation $0.08$, code match score variation $0.02$) proving substantially lower than the difference in average score between any two models. 

\noindent\textbf{Human Annotation and Powerful Judge:} The first author of the paper manually annotated description scores for a random subset of 50 predicted descriptions for each model and compared the implied ranking with the ranking of the judge model. For a subset of 25 of those predicted descriptions, we also wrote code snippets and used them to evaluate the code score. Additionally, we used GPT-5.5 (and Qwen3-Coder-Next for the coding model) to score /write code for the predicted descriptions and compared the implied rankings. In both cases, the rankings between models remained unchanged, showcasing the robustness of the judge and coding model.

\section{Interactive Mode Baseline:}
\label{sec:appendix_interactive}
We implement the \textbf{interactive} mode with the prompts given in Figure~\ref{sec:interactive_prompts}. Examples of the common failure modes are shown in Figure~\ref{fig:example}.
\begin{figure}
\begin{tcolorbox}
\textcolor{blue}{Conceptualization:}
You are given a Python function with the following header:
[HEADER]
Your task is to try various inputs to discover what this function does.

[CRITIQUE, empty unless we are running the RAG-based baseline]

So far, you have tried the following inputs: [PREV]
You then came up with the following running hypothesis: [HYPOTHESIS]

Based on this, what kind of input will you use to test the function with next? Very briefly describe your next intended input only, and the properties it satisfies. How does this input help test the hypothesis? What is the expected output? Be extremely concise and short. 
Your response should be extremely short and concise, just a few sentences. After the response, say [STOP]
Now provide your reasoning below and then say [STOP]
Reasoning:

\textcolor{blue}{Generation:}
You are given a Python function with the following header:
{header}
Your task is to try various inputs to discover what this function does.

So far, you have tried the following inputs: [PREV]
You then came up with the following running hypothesis: [HYPOTHESIS]

Based on this, you wanted to try the following kind of input next: [REASONING]. 
Now, give the exact input to test the function with next.
The input should be valid Python tuples and your output should follow the format below.
Suggested Input:
(arg0, arg1) [STOP] \#(arg0, arg1) should be replaced with actual input values in your response and must be a valid python tuple. This is an example format for a two arg function. You should adjust the number of arguments as per the function definition.
Now provide your suggested inputs below and then say [STOP]
Suggested Input:

\textcolor{blue}{Hypothesis Revision:}
You are given a Python function with the following header:
{header}
Your task is to try various inputs to discover what this function does.

So far, you have tried the following inputs: [PREV]
You then came up with the following running hypothesis: [HYPOTHESIS]
You wanted to test this, with an input coming from the reasoning: [REASONING]
Finally, you just tried the following inputs: [LAST\_INPUTS]

Based on this, can you conclude with very high confidence what the function does? If the function did not perform as you expected, the answer is likely no. If you think it is yes, then say YES and provide a concise description of its functionality.
Else, say NO and provide a revised hypothesis of what you think the function may do, and some guidance on how to test this further.
Format Example:
Hypothesis Conclusion: YES/NO
Summary: <your extremely concise summary or brief revised hypothesis here>
[STOP]

Now, provide your conclusion below, remember to say [STOP] after your summary.
Hypothesis Conclusion:
\end{tcolorbox}
\caption{Interactive mode prompts}
\label{sec:interactive_prompts}
\end{figure}

All models are evaluated with greedy sampling (however, we noticed persistent stochasticity in the responses of API-based models, despite setting the temperature to 0. Such results have been noted in other works~\citep{atil-etal-2025-non} and are likely a consequence of cost optimization on the side of model providers). Every call permits the model to generate up to 1000 tokens. In practice, this limit is rarely reached. 

Models are limited to a maximum of 25 API calls before they must confirm their hypothesis. 

\begin{figure}
\begin{tcolorbox}

\textcolor{blue}{True Description:} Removes all occurrences of the `+' character from the input 
string 'arg0' and returns the modified string.

\textbf{Contextual Examples}

("hello") => hello,

("Hello",) => Hello, 

\textbf{Explored Inputs}: 

("World",) => World 

("world",) => world, 

(42,) => Error: arg0 must be a string, 

("HELLO",) => HELLO, 

("test",) => test

\textcolor{red}{Predicted Description}: the function validates that arg0 is a string 
(raising an error if not) and returns the string unchanged. [Incorrect, misses `+' functionality]

\rule{\linewidth}{0.4pt}

\textcolor{blue}{True Description:} Checks if the input string is not a decimal number [decimal $\neq$ digit]. It returns True if the string contains non-decimal characters, otherwise it returns False.

\textbf{Contextual Examples:}

("abc",) => True, 

("123") => False, 

\textbf{Explored Inputs}:

("!@\#") => True, 

("a1",) => True, 

("",) => True, 

("½",) => True

\textcolor{red}{Predicted Description}: returns `not arg0.isdigit()` — it returns `true` if the string is not composed entirely of digit characters, and `false` if it is. [Incorrect, misses `.' functionality]
\end{tcolorbox}
\caption{Models (examples from Claude-4-Opus) exhibiting shallow exploration, and hence failing to reveal the full functionality of the API method. In both cases, testing more extensive edge cases and varied inputs could have revealed behaviour that contradicted the model's hypothesis. Instead, it queries the API with inputs that are similar to previous attempts, resulting in premature termination due to apparent confirmation of the hypothesis.}
\label{fig:example}
\end{figure}

We perform several experiments to verify that the interactive mode is a fair estimate of a model's ability to conduct interactive API discovery. We do this by implementing several alternative baseline approaches, as well as attempting to enhance the interactive mode with best practices from prompt engineering~\citep{sahoo2024systematic} and agentic memory with context engineering~\citep{mei2025survey}. Our results establish that these simpler changes to the method do not substantially improve performance, demonstrating that the interactive mode is a reasonable proxy for model ability.

Unless stated otherwise, all experiments below were conducted with the following models: Qwen3-1.7B, Qwen3-8B, GPT-4o-mini. 

\subsection{Prompt and Structure Variation}
For each of the prompts used, we wrote three variations (semantically equivalent) and hence deployed three variants of the interactive mode baseline. The variance of the average description score across models $(0.0061)$ was far smaller than the average description score difference between models. Qualitatively, models produced similar descriptions for similar test instances, with minor variations in the specific words used to describe them and the inputs used to explore the API.

We additionally tried two alternate structures of reasoning. 

\noindent\textbf{Batch Interactive Mode:} We implement a \textbf{batch interactive} variant that replaces the single-input-per-iteration loop with a breadth-first exploration strategy.
Rather than proposing one input and immediately reflecting on its result, the model is prompted to generate a small set of $B=3$ candidate inputs per iteration, each explicitly targeting a different candidate hypothesis about the function's behavior.
All inputs are executed against the black-box function, and the full set of results is passed to a single synthesis call that eliminates inconsistent hypotheses and produces a revised hypothesis.
This reduces the number of LLM calls per iteration from three to two while increasing the information gathered per round, potentially making it particularly effective early in exploration when the hypothesis space is wide and a single input is unlikely to discriminate between competing explanations.
The concluded/not-concluded decision is made at the synthesis step, and the loop terminates under the same conditions as the standard interactive baseline.

\noindent\textbf{Structured Hypothesis Tracking:} 
We implement a \textbf{structured interactive} variant that makes the model's belief state explicit throughout exploration.
Rather than maintaining a single free-form hypothesis, the model tracks a ranked list of $H \leq 5$ mechanistic hypotheses, each annotated with a confidence score and a summary of supporting and missing evidence.

Each iteration proceeds in three stages.
\begin{enumerate}
    \item The model is prompted to enumerate or revise its hypothesis list given all prior results, assigning confidence scores and identifying what evidence each hypothesis still requires to be confirmed or falsified.
    \item Given the full hypothesis list, the model designs $B=3$ inputs that maximally discriminate between competing hypotheses, prioritizing edge cases and boundary conditions. All three inputs are executed against the black-box function.
    \item Third, the model updates its hypothesis list in light of the new results, eliminating hypotheses inconsistent with the observed outputs and revising confidence scores accordingly.
\end{enumerate}First, 
Exploration terminates when the model either explicitly concludes with high confidence or only a single hypothesis survives with no simple falsifying test remaining.
This structure enforces falsificationist discipline: each proposed input must be justified by which hypotheses it distinguishes, preventing the model from repeatedly probing variations of already-confirmed behavior.

\noindent\textbf{Results:} For both of these variations, we notice considerable \textbf{performance degradations} compared to the default interactive baseline for the Qwen3-1.7B and Qwen3-8B models. GPT-4o-mini achieved a similar score (2.419 vs 2.384) with the batch setting. However, it performs worse in the structured hypothesis tracking mode (2.12). A qualitative inspection revealed that creating more sophisticated prompt scaffold pipelines with complex expectations of the model required more brittle output parsing and led to long context buildups, which eventually degraded model performance. 

\subsection{Injecting Software Testing Specific Knowledge}
Having seen evidence that more complex workflows would compromise model performance, we asked whether models could meaningfully improve when provided with software testing-specific guidance in the prompt. We replaced the `[CRITIQUE]` section of the reasoning prompt with the following text, drawing from best practices in black box software testing~\citep{myers2004art, kery2017exploring}:

\begin{verbatim}
PARTITION THE INPUT SPACE FIRST. Do not probe random variations — 
identify the equivalence classes of the function's domain and 
ensure you have at least one example from each. 
Typical partitions: empty/null inputs, singleton, small valid, large valid, 
boundary values, invalid types, negative/zero/maximum values.

USE CONTRASTING PAIRS. To isolate what a single argument controls, vary exactly 
one dimension at a time while holding all others fixed. 
If f(A, B) → X and f(A', B) → X, then B is likely irrelevant to that behavior. 
This is the only reliable way to attribute 
behavior to a specific argument.

TEST THE BOUNDARIES BEFORE THE MIDDLE. The most informative inputs are: the smallest 
valid input, the largest valid input, inputs just inside and just outside 
any apparent threshold, and the degenerate cases 
(empty list, zero, empty string, None). 
These reveal the function's contract faster than any mid-range probe.

PREDICT BEFORE YOU PROBE. Before submitting an input, state what 
you expect the output to be. 
If the actual output surprises you, your hypothesis is wrong — 
do not rationalize it. Discard the hypothesis and form a new one.

AVOID REDUNDANCY. If a new input would produce the same output 
regardless of which of your current candidate hypotheses is true, 
it is uninformative. 
Choose inputs that produce different outputs under different hypotheses.

DISTINGUISH SIMILAR BEHAVIORS AGGRESSIVELY. Common confusions: 
(is it sorting or filtering?), 
(is it checking membership or counting?), (is it a threshold or a modulo?), 
(is it working on the values or the indices?). Design inputs specifically 
to break these symmetries.

DO NOT OVER-EXPLAIN NORMAL BEHAVIOR. Once a case is confirmed, stop testing it. 
Spend your remaining budget on unverified branches and failure modes.
\end{verbatim}

\noindent\textbf{Results:} The addition of this prompt has an overall negative affect across all models evaluated (Qwen3-1.7B, Llama3-8B, Qwen3-8B, GPT-4o-mini). We also confirm this with evaluations of other models on a subset of the test set (Gemini-3-Flash, Claude-4-Opus). Qualitatively, the smaller models break down due to an increase in context length, while the more powerful systems seem to excessively focus on `similar looking pairs' and do not engage in wider exploration of the API's functionality.

\subsection{RAG-based approach}
Finally, we considered whether the limitation of the previous approach was that the prompt was too generic, with little guidance about how to best explore the particular test method at hand. We attempt to create a more sophisticated, memory-based context engineering system that can potentially serve the model with specific guidance on how to best explore the test API in question. 

To do so, we implement a \textbf{memory} baseline that augments the interactive mode with a retrieve-then-reflect mechanism inspired by episodic memory.
The method proceeds in three stages.

\noindent\textbf{Stage 1: Initial Exploration.}
The model first runs the interactive baseline on the test function, producing an initial hypothesis and a record of all input--output pairs it queried.

\noindent\textbf{Stage 2: Retrieval.}
The initial hypothesis is embedded using \textsc{Qwen3-Embedding-8B}~\citep{zhang2025qwen3} and compared against pre-computed embeddings of all training-set function descriptions via cosine similarity.
The $k=5$ most similar training functions are retrieved.

\noindent\textbf{Stage 3: Critique and Re-exploration.}
For each retrieved training function---whose ground-truth description is known---the model re-runs the interactive exploration procedure and then generates a self-critique: it is shown its predicted description alongside the true description and the examples it chose to query versus examples that would have revealed the true behavior, and is asked to identify gaps in its exploration strategy.
When multiple critiques are produced, they are consolidated into a single unified critique by a summarization call.
The model then runs the interactive procedure a second time on the original test function, initialised with its prior hypothesis and queried examples, and with the consolidated critique prepended to the reasoning prompt as a piece of additional guidance.
This potentially allows the model to correct systematic blind spots in its exploration strategy identified from analogous functions in the training set.

\noindent\textbf{Result:}
We did not observe a significant increase in performance when using additional memory. The performance of GPT-4o-mini and Qwen3-1.7B decreases marginally, as they end up incorrectly revising accurate hypotheses after the critiques are provided. Qwen3-8B improves marginally ($+0.015$), however, this is not statistically significant at $p<0.2$.

\subsection{Incontext Baseline}
As described in Section~\ref{sec:results}, we use the \textbf{incontext} mode to quantify the extent to which access to an API improves a model's understanding of its functionality. 

In particular, we use a similar prompt to the final step of the \textbf{interactive} mode (Figure~\ref{sec:incontext_prompts}), providing only the two contextual examples and requesting a direct prediction of the description. 

\begin{figure}
\begin{tcolorbox}
You are given a Python function with the following header:
def test\_func(arg0, arg1):
You have tried the following inputs to discover what this function does.

[
Input: ("input test", "+") => Output: input+test, Error: None
Input: ("", "-") => Output: '', Error: None
]

Based on this, describe what the function does in words. Respond in the format: Description: [your description here] [STOP] \# make sure to put [STOP] after your description.
Now, provide your description for the function:
Description:
\end{tcolorbox}
\caption{Incontext mode prompt example}
\label{sec:incontext_prompts}
\end{figure}

\begin{figure}
    \centering
    \includegraphics[width=0.75\linewidth]{images/description_score_spread.pdf}
    \includegraphics[width=0.75\linewidth]{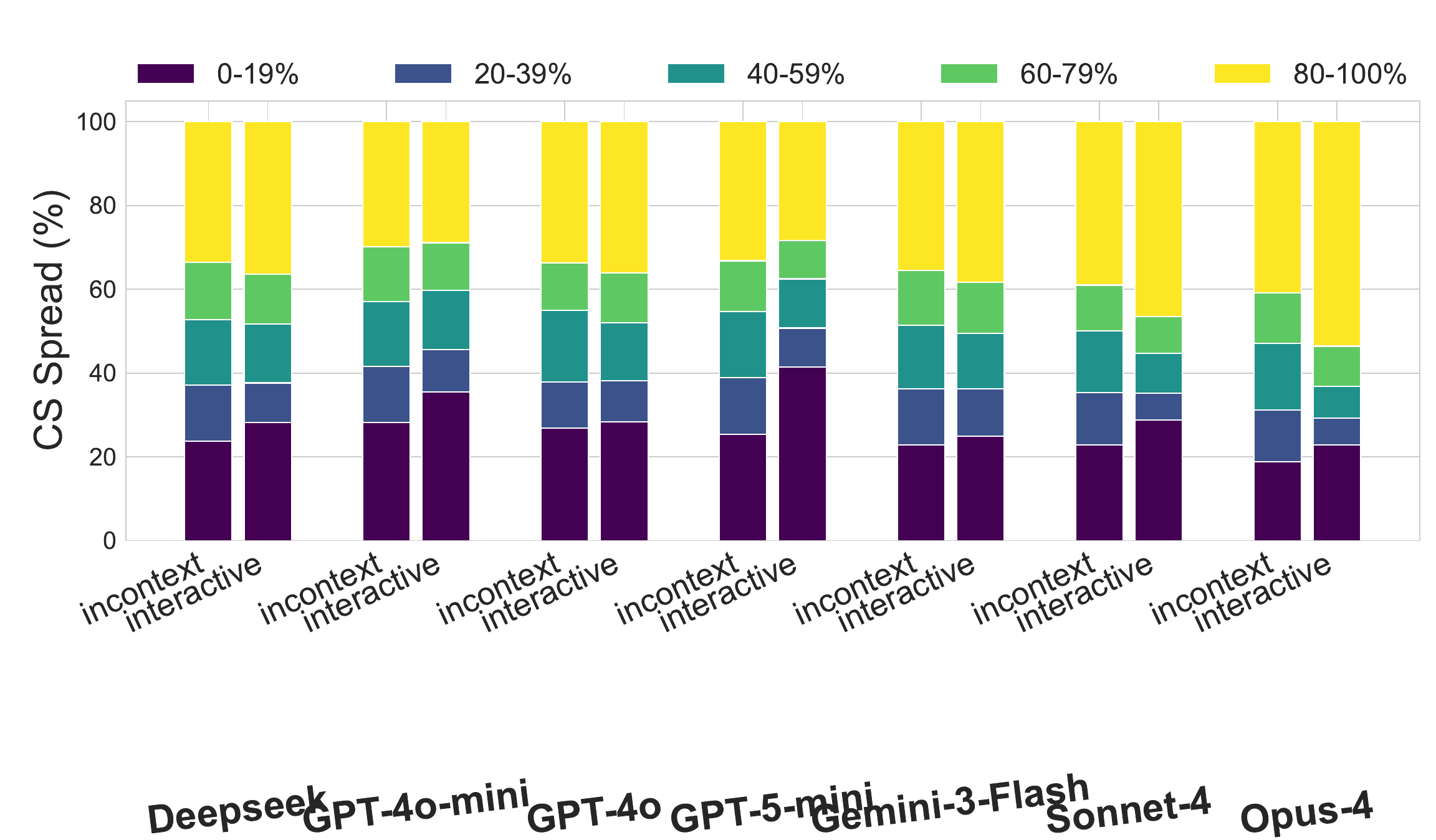}
    \caption{Comparison between the incontext and interactive setting shows that models struggle to fully utilize the benefits of interactive API access.}
    \label{fig:incontext_results}
\end{figure}

For the majority of models evaluated, performance in this setting is not significantly lower than the performance in the \textbf{interactive} mode (at $p<0.05$ with a bootstrap t-test). We show results for the best performing models (Figure~\ref{fig:incontext_results}), establishing that even frontier models like Gemini-3-Flash struggle to use their interactive access to the fullest. 

\section{Asymmetric Actor Critic Details}
\label{sec:appendix_aac}
\subsection{Training Hyperparameters}

We train using GRPO~\cite{DeepSeekAI2025DeepSeekR1IR} via the SkyRL framework with the hyperparameters listed in Table~\ref{tab:hyperparams}.

\begin{table}[h]
\centering
\caption{SkyRL training hyperparameters.}
\label{tab:hyperparams}
\begin{tabular}{lc}
\toprule
\textbf{Hyperparameter} & \textbf{Value} \\
\midrule
\multicolumn{2}{l}{\textit{Model}} \\
Model dtype & \texttt{bfloat16} \\
LoRA rank & 16 \\
LoRA alpha & 16 \\
\midrule
\multicolumn{2}{l}{\textit{Training}} \\
Epochs & 5 \\
Update epochs per batch & 1 \\
Train batch size & 8 \\
Policy mini-batch size & 8 \\
Micro train batch size per GPU & 2 \\
Policy optimizer learning rate & $1\times10^{-6}$ \\
Advantage estimator & GRPO \\
KL loss & enabled \\
Strategy & FSDP2 \\
\midrule
\multicolumn{2}{l}{\textit{Generation}} \\
Samples per prompt ($G$) & 8 \\
Max prompt length & 512 tokens \\
Max generation length & 1024 tokens \\
Max model context length & 4096 tokens \\
Generator backend & vLLM \\
GPU memory utilization & 0.8 \\
\midrule
\multicolumn{2}{l}{\textit{Environment}} \\
Max turns per episode & 40 \\
Function execution timeout & 10 s \\
\midrule
\multicolumn{2}{l}{\textit{Checkpointing}} \\
Checkpoint interval & 10 steps \\
Eval interval & 5 steps \\
\bottomrule
\end{tabular}
\end{table}

\subsection{Reward Schedule (Supervised Setting)}

Each episode follows a fixed three-step cycle: \textbf{reasoning} $\to$ \textbf{input proposal} $\to$ \textbf{reflection}. Each step receives an independent scalar reward. We use three global constants:
\[
  \lambda_H = 2.0, \quad p_{\text{parse}} = 0.5, \quad p_{\text{len}}^{\max} = 1.0.
\]

\paragraph{Length penalty.}
All three steps include a length penalty:
\begin{equation}
  r_{\text{len}}(x;\, \tau,\, \alpha) =
  \begin{cases}
    0 & |x|_{\text{tok}} \leq \tau \\[4pt]
    -\min\!\left(\dfrac{\alpha\,(|x|_{\text{tok}} - \tau)}{2\,\alpha\,\tau},\; p_{\text{len}}^{\max}\right) & \text{otherwise.}
  \end{cases}
\end{equation}

\paragraph{Step 1 — Reasoning.}
The actor produces a brief rationale for which input to try next.
\begin{equation}
  r_{\text{reasoning}} = r_{\text{judge}}^{\text{reason}} + r_{\text{len}}(\text{response};\; \tau{=}100,\, \alpha{=}0.05)
\end{equation}
where $r_{\text{judge}}^{\text{reason}} \in [0,1]$ is the LLM judge's rating (0--9 scale, divided by 9) of how informative the proposed input is for discriminating hypotheses.

\paragraph{Step 2 — Input Proposal.}
The actor proposes the exact Python argument tuple to pass to the black-box function.
\begin{equation}
  r_{\text{input}} =
  \begin{cases}
    -p_{\text{parse}} = -0.5 & \text{parse failure} \\[4pt]
    +\dfrac{\lambda_H}{2\,T_{\max}} = +0.025 & \text{clean execution} \\[4pt]
    -p_{\text{parse}} = -0.5 & \text{runtime error}
  \end{cases}
  \;+\; r_{\text{len}}(\text{input};\; \tau{=}30,\, \alpha{=}0.5)
\end{equation}
The execution bonus is intentionally small ($0.025$) so it never competes with the hypothesis reward. The tight length threshold ($\tau{=}30$) penalizes verbose inputs.

\paragraph{Step 3 — Reflection (Hypothesis Update).}
The actor observes the execution result, declares \textsc{yes}/\textsc{no} on whether it has identified the function, and states its current hypothesis.
\begin{equation}
  r_{\text{reflect}} = \mathbb{1}[\text{parse fail}]\cdot(-p_{\text{parse}}) + r_{\text{hyp}}(\text{done}) + r_{\text{len}}(\text{response};\; \tau{=}100,\, \alpha{=}0.05)
\end{equation}
\begin{equation}
  r_{\text{hyp}}(\text{done}) = \lambda_H \cdot \bigl(s_{\text{hyp}} + b_{\text{term}}(\text{done},\, s_{\text{hyp}})\bigr)
\end{equation}
where $s_{\text{hyp}} = \text{judge\_rating}/9 \in [0,1]$ is the LLM judge's score of the current hypothesis against the ground-truth description, and
\begin{equation}
  b_{\text{term}}(\text{done},\, s) =
  \begin{cases}
    +1.0 & \text{done} = \textsc{true},\; s > 0.7 \quad\text{(correct termination)} \\
    -1.0 & \text{done} = \textsc{true},\; s \leq 0.7 \quad\text{(premature termination)} \\
    +1.0 & \text{done} = \textsc{false},\; s < 0.3 \quad\text{(correctly continuing)} \\
    0    & \text{otherwise.}
  \end{cases}
\end{equation}

\paragraph{Timeout.}
If the episode reaches $T_{\max} = 40$ steps, a terminal penalty of $-1.0$ is applied.

\paragraph{Summary.}
Table~\ref{tab:rewards} enumerates every reward signal and its range.

\begin{table}[h]
\centering
\caption{Complete reward schedule in the supervised setting.}
\label{tab:rewards}
\begin{tabular}{llcc}
\toprule
\textbf{Step} & \textbf{Signal} & \textbf{Value / Range} & \textbf{Source} \\
\midrule
\multirow{2}{*}{Reasoning}
  & Judge score (reasoning quality) & $[0,\,1]$        & LLM judge ($\div\,9$) \\
  & Length penalty                  & $[-1,\,0]$       & $\tau{=}100$, $\alpha{=}0.05$ \\
\midrule
\multirow{4}{*}{Input proposal}
  & Parse failure                   & $-0.5$           & hard rule \\
  & Clean execution                 & $+0.025$         & $\lambda_H/(2T_{\max})$ \\
  & Runtime error                   & $-0.5$           & hard rule \\
  & Length penalty                  & $[-1,\,0]$       & $\tau{=}30$, $\alpha{=}0.5$ \\
\midrule
\multirow{4}{*}{Reflection}
  & Parse failure                   & $-0.5$           & hard rule \\
  & Hypothesis score ($s_{\text{hyp}}$) & $[0,\,1]$   & LLM judge ($\div\,9$) \\
  & Termination bonus               & $\{-1,\,0,\,+1\}$ & threshold on $s_{\text{hyp}}$ \\
  & Hypothesis scale                & $\times\,2.0$    & $\lambda_H$ applied to above \\
  & Length penalty                  & $[-1,\,0]$       & $\tau{=}100$, $\alpha{=}0.05$ \\
\midrule
Episode & Timeout penalty           & $-1.0$           & at $T{=}40$ turns \\
\bottomrule
\end{tabular}
\end{table}

\subsection{What Does the Asymmetric Critic Add?}
\label{sec:appendix_aac_baselines}

To isolate the contribution of each component of AAC, we compare against four alternate training paradigms (Table~\ref{tab:rl_baselines}):

\noindent\textbf{DISTILL:} Supervised fine-tuning on interactive trajectories generated by a more powerful model (GPT-4o-mini). This distils basic capabilities without RL.

\noindent\textbf{CodeARC-SFT:} Following \citet{wei2025codearc}, we generate trajectories on the \benchmarkshort{} training set using the base model as a `teacher' that is given the true function. We instruct the teacher to reason as if it were deriving the logic from its own intuition, without explicitly referring to the true function, and perform supervised fine-tuning on these trajectories (including reasoning) using prompts that do not reveal the true function.

\noindent\textbf{FINAL:} AAC without the intermediate, step-level rewards, i.e. the only reward is the LM-as-a-judge score of the hypothesis when the agent decides to terminate.

\noindent\textbf{SYM-INFO:} GRPO with the same reward schedule as AAC, but with rewards granted by a critic that does not see the ground truth, i.e. a symmetric actor critic.

\begin{table}[h]
\centering
\caption{Description score (LM-as-a-judge, 1--5) of AAC against alternate training paradigms. AAC is the only approach that improves substantially over the base model. Training on trajectories from an answer-aware teacher (CodeARC-SFT) or rewarding with a critic that lacks privileged information (SYM-INFO) actively harms performance.}
\label{tab:rl_baselines}
\begin{tabular}{lcc}
\toprule
\textbf{Method} & \textbf{Qwen3-1.7B} & \textbf{Llama3-8B} \\
\midrule
Base & 1.92 & 1.99 \\
\midrule
DISTILL & 1.96 & 2.02 \\
CodeARC-SFT & 1.81 & 1.79 \\
FINAL & 1.86 & 1.91 \\
SYM-INFO & 1.22 & 1.34 \\
\midrule
AAC & \textbf{2.04} & \textbf{2.13} \\
\bottomrule
\end{tabular}
\end{table}

AAC outperforms all alternatives, with each comparison isolating a distinct component of its success. SYM-INFO is particularly harmful, as the critic must judge hypotheses and exploratory inputs in an ungrounded manner, showing that the \textbf{privileged information} is vital. While FINAL uses only sound reward signals, we find that its loss does not converge when the signal is so sparse, showing that \textbf{dense, step-wise guidance} is necessary. DISTILL yields small gains, but falls short of AAC despite relying on a more powerful model for guidance.

Most notably, CodeARC-SFT degrades performance below the base model, despite leveraging the same privileged information as AAC. When the teacher is aware of the true function, it rarely takes more than three queries to fully expose the API's behaviour, and will often use only one query before terminating (the fine-tuned models make an average of $0.13$ and $0.57$ queries, respectively, compared to $12.50$ and $15.27$ after AAC-tuning). Qualitatively, the teacher queries exactly the boundary conditions of the function before concluding with a hypothesis. This problem goes deeper than the trajectories being off-policy: the process followed by a model that knows the true function is fundamentally different from the process of discovering it. An agent exploring an unknown function must be more exploratory and less precise, trying inputs that may yield surprising results and refine its hypothesis. The teacher trajectories are hence `too optimal', and training on them biases the model towards making fewer queries and exploring the API less thoroughly. In contrast, AAC uses the privileged information only for reward assignment, while the text sequences on which the loss is optimised are always generated by the agent itself. This division of roles allows us to train on realistic exploration trajectories, while rewarding them with the benefit of privileged knowledge.

\section{Computational Hardware}
The inference for most of our experiments was run on a compute cluster with 8 NVIDIA A40 GPUs (approx 46068 MiB of memory) on CUDA version 12.6. The CPU on the cluster is an AMD EPYC 7502 32-Core Processor. Most experiments could be conducted with less than 16GB of GPU RAM. 

AAC training of 8B models was run on a separate system with 2 NVIDIA H100 GPUs. 

Benchmark test run times vary drastically with the model used, ranging from 3 hours to 120 hours. 

For several large open source models (e.g. DeepSeek), we use OpenRouter for inference. 

\newpage

\end{document}